\def\year{2021}\relax
\documentclass[letterpaper]{article} %
\usepackage{aaai21}  %
\usepackage{times}  %
\usepackage{helvet} %
\usepackage{courier}  %
\usepackage[hyphens]{url}  %
\usepackage{graphicx} %
\usepackage{natbib}  %
\usepackage{caption} %
\usepackage[utf8]{inputenc} %
\usepackage[T1]{fontenc}    %
\usepackage[hidelinks]{hyperref} %
\usepackage{booktabs}[color]       %
\usepackage[utf8]{inputenc} %
\usepackage[T1]{fontenc}    %
\usepackage{url}            %

\usepackage{amsfonts}       %
\usepackage{nicefrac}       %
\usepackage{microtype}      %
\usepackage{textcomp}
\usepackage{siunitx}
\usepackage{comment}
\usepackage{multirow}
\usepackage{mathtools}

\usepackage{filecontents} %

\usepackage{algpseudocode}
\usepackage{makecell}

\usepackage{enumerate}[leftmargin=*]
\usepackage{enumitem,amsmath,amssymb,bbm}

\usepackage[linesnumbered, ruled]{algorithm2e}  %

\SetCommentSty{mycommfont}
\SetAlFnt{\footnotesize}

\usepackage{minitoc}  %

\usepackage{textcomp}
\usepackage{siunitx}
\usepackage{graphicx}  %
\usepackage{comment}
\usepackage{multirow}
\usepackage{booktabs} %
\usepackage{filecontents} %
\usepackage{amsmath}
\usepackage{amsfonts}
\usepackage{array}
\usepackage{bm}
\usepackage{caption,subcaption}
\usepackage{pgfplots}
\usepackage{pgfplotstable}
\usepackage{xcolor}

\newcommand{\T}[1]{\ensuremath{\mathcal{#1}}} %
\newcommand{\M}[1]{\ensuremath{\mathbf{#1}}} %
\newcommand{\V}[1]{\ensuremath{\mathbf{#1}}} %

\usepackage{url}
\usepackage{xspace,csquotes}
\usepackage{enumitem}

\usepackage{stmaryrd}
\newtheorem{definition}{Definition}
\newtheorem{proposition}{Proposition}

\newcommand{\mname}{\texttt{SWIFT}\xspace}

\newcommand\norm[1]{\left\lVert#1\right\rVert}

\title{\mname: Scalable Wasserstein Factorization for Sparse Nonnegative Tensors}
\author{
    Ardavan Afshar,\textsuperscript{\rm 1} 
    Kejing Yin,\textsuperscript{\rm 2} 
    Sherry Yan,\textsuperscript{\rm 3} 
    Cheng Qian,\textsuperscript{\rm 4} 
    Joyce Ho,\textsuperscript{\rm 5} 
    Haesun Park,\textsuperscript{\rm 1} 
    Jimeng Sun\textsuperscript{\rm 6} \\
}
\affiliations{
    \textsuperscript{\rm 1} Georgia Institute of Technology \textsuperscript{\rm 2} Hong Kong Baptist University \textsuperscript{\rm 3} Sutter Health 
    \textsuperscript{\rm 4} IQVIA \\
    \textsuperscript{\rm 5} Emory University 
    \textsuperscript{\rm 6} University of Illinois at Urbana-Champaign \\
    aafshar8@gatech.edu, cskjyin@comp.hkbu.edu.hk, YanSX@sutterhealth.org, alextoqc@gmail.com, joyce.c.ho@emory.edu, hpark@cc.gatech.edu, jimeng@illinois.edu
}

\begin{document}

\doparttoc %
\faketableofcontents %
\parttoc %

\maketitle

\begin{abstract}
Existing tensor factorization methods assume that the input tensor follows some specific distribution (i.e. Poisson, Bernoulli, and Gaussian), and solve the factorization by minimizing some empirical loss functions defined based on the corresponding distribution. However, it suffers from several drawbacks:  1) In reality, the underlying distributions are complicated and unknown, making it infeasible to be approximated by a simple distribution. 2) The correlation across dimensions of the input tensor is not well utilized, leading to sub-optimal performance. Although heuristics were proposed to incorporate such correlation as side information under Gaussian distribution, they can not easily be generalized to other distributions. Thus, a more principled way of utilizing the correlation in tensor factorization models is still an open challenge. Without assuming any explicit distribution, we formulate the tensor factorization as an {\it optimal transport} problem with Wasserstein distance, which can handle non-negative inputs. 
 
We introduce \mname, which minimizes the Wasserstein distance that measures the distance between  the input tensor and that of the reconstruction. In particular, we define the $N$-th order tensor Wasserstein loss for the widely used tensor CP factorization and derive the optimization algorithm that minimizes it. By leveraging sparsity structure and  different equivalent formulations for optimizing computational efficiency, \mname is as scalable as other well-known CP algorithms. Using the factor matrices as features, \mname achieves up to 9.65\% and  11.31\% relative improvement over baselines for  downstream prediction tasks. Under the noisy conditions, \mname achieves up to $15\%$ and $17\%$ relative improvements over the best competitors for the prediction tasks. 

\end{abstract}

\section{Introduction}\label{sec:introduction}

Tensor factorization techniques are  effective and powerful tools for analyzing multi-modal data and  have been shown tremendous success  in a wide range of applications including spatio-temporal analysis \cite{bahadori2014fast,fanaee2016event}, graph analysis \cite{gujral2020beyond},  and health informatics \cite{yin2019learning, he2019distributed} applications. Many constraints such as non-negativity \cite{kim2014algorithms}, sparsity \cite{henderson2017granite}, orthogonality \cite{wang2015rubik}, and smoothness \cite{afshar2018copa} are imposed on tensor methods in order to improve the performance both quantitatively and qualitatively. Moreover, depending on the nature of input data, tensor methods fit different distributions on data including Gaussian \cite{bader2007efficient}, Poisson \cite{chi2012tensors}, Bernoulli \cite{hong2020generalized} distributions and minimize various empirical loss functions such as sum square loss, KL-divergence, and log-loss. However, there are several limitations with these techniques. 1) Existing factorization models often assume some specific data distributions, yet in practice, the underlying distributions are complicated and often unknown.  2) The nature of these factorization models neglects correlation relations within each tensor mode (such as external knowledge about similarity among those features). Although there are several extensions to tensor factorization approaches  that consider these similarity matrices as side information \cite{acar2011all, kim2017discriminative}, they are derived under Gaussian distribution and are not directly generalizable to unknown distributions.

Recent success of {\it Wasserstein distance or loss} (\textit{a.k.a.} \textit{earth mover's distance} or {\it optimal transport} distance) shows its potential as a better measure of the difference between two distributions \cite{WGAN, cuturi2013sinkhorn,frogner2015learning}. This distance metric provides a natural measure of the distance between two distributions via a ground metric of choice and can be defined as the cost of the optimal transport plan for moving the mass in the source distribution to match that in the target one. Recently,  Wasserstein distance has been applied to matrix factorization and dictionary learning problems with great success \cite{sandler2009nonnegative, rolet2016fast, qian2016non, schmitz2018wasserstein, varol2019temporal}. To the best of our knowledge, its extension to tensor factorization was never studied and is in fact non-trivial due to the following challenges: 

\begin{itemize}[leftmargin=*]
    \item \textbf{Wasserstein loss is not well-defined for tensors:} The Wasserstein loss is originally defined over vectors, where each entry of the vector represents one physical location and the cost of transporting from one location to another is used. Existing matrix-based Wasserstein loss are defined by the sum of the vector-based Wasserstein loss over the columns of the matrices \cite{rolet2016fast}. 
    Unfortunately, this definition is not applicable to tensors of multiple modes. %
    \item \textbf{Wasserstein loss is difficult to scale:} Learning with Wasserstein loss generally requires solving the optimal transport problem in each iteration, which is extremely time-consuming. %
    \item \textbf{Large and sparse nonnegative input:} Existing works on Wasserstein matrix factorization are developed based on dense input with relatively small size. In reality, tensors can often be very large are are often sparse. Efficient learning algorithms are possible only when the sparsity structures of the large input tensor are properly identified and fully utilized.  %
    
\end{itemize}

To overcome these challenges,  we propose \mname, a tensor factorization method which efficiently minimizes  Wasserstein distance for sparse nonnegative tensors. 
The main contributions of \mname  include:

\begin{itemize}[leftmargin=*]   
    \item \textbf{Defining Optimal Transport for Tensors that Handles Nonnegative Input:} \mname is the first technique that minimizes optimal transport (OT) distance for tensor factorization approaches.  The benefit of OT  is that it does not assume any specific distribution in the data.  \mname is able to handle nonnegative inputs such as binary, counts and probability measures, which are common input in real-world tensor data. %
    \item \textbf{Full Utilization of Data Sparsity and Parallelism:}  By fully exploring and utilizing the sparsity structure of the input data, \mname significantly reduces the number of times required to compute OT and enables parallelism. \mname obtains up  to $16 \times$ faster OT computation than direct implementation of Wasserstein tensor factorization.%
    \item \textbf{Efficient Computation:} \mname reduces the amount of computations by smartly rearranging the objective function for solving each factor matrix. Scalability of \mname is comparable with well-known CP algorithms. Moreover, \mname achieves up to $921 \times$ speed up over a direct implementation of Wasserstein tensor factorization without our speedup strategies, as shown in the  appendix.  
\end{itemize}

\section{Notations and Background}

\subsection{Basic Notations and Tensor Operations}
We denote vectors by bold lowercase letters (\textit{e.g.} \V{u}), matrices by bold uppercase letters (\textit{e.g.} \M{A}), and tensors by Euler script letters  (\textit{e.g.} $\T{X}$).
The entropy $E$ for a nonnegative matrix $\M{A} \in \mathbb{R}_{+}^{M \times N}$ is defined as $E(\M{A})= -\sum_{i,j=1}^{M,N} \M{A}(i,j) \text{log}(\M{A}(i,j)) $.
$KL(\M{A}||\M{B})$ is the generalized KL-divergence between two matrices $\M{A}, \M{B} \in \mathbb{R}^{M \times N}$  is defined as  $KL(\M{A}||\M{B})= \sum_{i,j=1}^{M,N} \M{A}(i,j) \text{log}(\frac{ \M{A}(i,j)}{ \M{B}(i,j)})- \M{A}(i,j)+\M{B}(i,j)$. 

\textbf{Mode-$\bm{n}$ Matricization.} Matricization~\cite{kolda2009tensor} is the process of reordering the entries of a tensor into a matrix. Specifically, the mode-$n$ matricization is the concatenation of all the mode-$n$ fibers obtained by fixing the indices for every but the $n^{th}$ mode. It transforms the tensor $\T{X} \in \mathbb{R}^{I_1 \times I_2 ... \times I_N} $ into matrix $\M{X_{(n)}}$, and the size of the resulting matrix is $I_n$ by  $I_1 ...I_{n-1}I_{n+1}... I_N$. To ease the notation, we define $I_{(-n)} =I_1 ...I_{n-1}I_{n+1}... I_N$.

\textbf{Khatri-Rao Product.} The Khatri-Rao product~\cite{kolda2009tensor} of two matrices $\mathbf{A}\in\mathbb{R}^{I\times R}$ and $\mathbf{B}\in\mathbb{R}^{J\times R}$ is the column-wise Kronecker product $\mathbf{C} = \mathbf{A} \odot \mathbf{B}=\left[
\mathbf{a}_{1}  \otimes  \mathbf{b}_{1} \quad  \mathbf{a}_{2}  \otimes \mathbf{b}_{2} \quad \cdots \quad \mathbf{a}_{R} \otimes \mathbf{b}_{R}\right],$ where $\mathbf{a}_i, \mathbf{b}_i$ are the column-$i$ of matrices $\mathbf{A}$ and $\mathbf{B}$, $\otimes$ denotes the Kronecker product, and $\mathbf{C}\in\mathbb{R}^{IJ\times K}$. We denote the Khatri-Rao product of all factor matrices except the $n$-th mode as
\begin{equation}\small
    \M{A}_{\odot}^{(-n)} = (\M{A}_N \odot ... \odot \M{A}_{n+1}\odot \M{A}_{n-1}\odot ...  \odot \M{A}_1) \in \mathbb{R}^{I_{(-n)} \times R},\label{eq:khatri_rao}
\end{equation}
where $\M{A}_{n} \in\mathbb{R}^{I_n\times R}$  indicates the $n$-th factor matrix. 

\textbf{Canonical/Polyadic (CP) decomposition.} The CP factorization~\cite{kolda2009tensor} approximates a tensor $\T{X}$ as the sum of rank-one tensors ( $ \hat{\T{X}}=\llbracket \M{A}^{(1)},\M{A}^{(2)},....,\M{A}^{(N)} \rrbracket=\sum_{r=1}^{R}\V{a}^{(1)}_r \circ \V{a}^{(2)}_r \circ ... \circ \V{a}^{(N)}_r$), where $\hat{\T{X}}$ is a reconstructed tensor,  $\V{a}^{(n)}_r$ is the $r$-th column of factor matrix $\M{A}^{(n)}$, $\circ$ denotes the outer product of vectors, and $R$ is the number of the rank-one tensors to approximate the input tensor, \textit{i.e.}, the target rank. The mode-n matricization of reconstructed tensor $\hat{\T{X}}$ is  $ \hat{\M{X}}_{(n)} =\M{A}_n(\M{A}_{\odot}^{(-n)})^T  \in \mathbb{R}^{I_n \times I_{(-n)}}$. %

\subsection{Preliminaries \& Related Work} \label{background}
\textbf{Wasserstein Distance and Optimal Transport.}
Wasserstein distance (\textit{a.k.a.} earth mover's distance or optimal transport distance) 
computes the  distance between two probability vectors\footnote{Vector \V{a} is a probability vector if $\norm{\V{a}}_1=1$ and all elements in $\V{a}$ are non-negative. }. Given two vectors $\V{a} \in \mathbb{R}_{ + }^{n}$, $\V{b} \in \mathbb{R}_{ + }^{m}$ and cost matrix $\M{C} \in \mathbb{R}_{ + }^{n \times m} $, the Wasserstein distance between \V{a}, \V{b} is shown by $W(\V{a}, \V{b})$ and  minimizes $\langle \M{C},\M{T} \rangle$  where $\langle .,. \rangle$ indicates the Frobenius inner product and $\M{C} \in \mathbb{R}_{+}^{n \times m} $ is a symmetric input cost matrix where  $\M{C}(i,j)$ represents the cost of moving $\V{a}[i]$ to $\V{b}[j]$.   $\M{T} \in U(\V{a}, \V{b}) $ where $\M{T}$ is an optimal transport solution between probability vectors \V{a} and \V{b} and  $U(\V{a}, \V{b})=\{\M{T} \in \mathbb{R}_{ + }^{n \times m} | \M{T}\V{1}_m=\V{a}, \M{T}^T\V{1}_n=\V{b}  \}$  is a set of all non-negative $n \times m$ matrices with row and column sums \V{a}, \V{b} respectively.  $\V{1}_m$ represents $m$ dimensional vector of ones. The aforementioned problem has complexity $O(n^3)$ (assuming $m=n$)~\cite{peyre2019computational}.  
However, computing this  distance metric comes with a heavy computational price \cite{pele2009fast}.  In order to reduce the complexity of computation, Cuturi et.al. \cite{cuturi2013sinkhorn} propose an entropy regularized optimal transport problem between  vectors \V{a}, \V{b}:
\begin{equation}
\small
\begin{aligned}
 W_V(\V{a}, \V{b})= \underset{\M{T} \in U(\V{a}, \V{b})}{\text{minimize}} \quad\langle \M{C},\M{T} \rangle - \frac{1}{\rho} E(\M{T}),
\end{aligned}
\label{Wass_def_regularized_entropy}
\end{equation}
where $E(\M{T})$ is an entropy function and $\rho$ is a regularization parameter. When $\rho \ge 0$, the solution of \eqref{Wass_def_regularized_entropy} is called the \textit{Sinkhorn divergence} (\textit{a.k.a.} entropy regularized Wasserstein distance) between probability vectors \V{a} and \V{b}. Eq.~\eqref{Wass_def_regularized_entropy} is a strictly convex problem with a unique solution and can be computed with vectors $\V{u} \in \mathbb{R}_{+}^{n}, \V{v} \in \mathbb{R}_{+}^{m}$ such that $\operatorname{diag}(\V{u})\M{K} \operatorname{diag}(\V{v}) \in U(\V{a}, \V{b}) $. Here, $\M{K} =\text{exp} (-\rho\M{C}) \in \mathbb{R}_{+}^{n \times m}$. Finding the optimal $\V{u}$ and $\V{v}$  can be computed via the Sinkhorn's algorithm~\cite{sinkhorn1967concerning}.%

\textbf{Wasserstein Dictionary Learning.} There are several techniques for  minimizing Wasserstein loss for dictionary learning problem \cite{sandler2009nonnegative, rolet2016fast, qian2016non,schmitz2018wasserstein,varol2019temporal,xu2020gromov}. Sandler et.al. \cite{sandler2009nonnegative} introduces the first non-negative matrix factorization problem that minimizes Wasserstein loss by proposing a linear programming problem, however, their method needs heavy computation. Cuturi \cite{rolet2016fast} proposed a Wasserstein dictionary learning problem based on entropy regularization. %
\cite{qian2016non} proposes a similar method by exploiting knowledge in both data manifold and features correlation. Xu~\cite{xu2020gromov} introduces a  nonlinear matrix factorization approach for graphs that considers topological structures. Unfortunately, these approaches cannot be directly generalized to tensor inputs due to the challenges mentioned in Section~\ref{sec:introduction}.

\section{\mname Framework} \label{swift_framwork}
We define and solve optimal transport problem for tensor input by proposing \underline{S}calable \underline{W}asserste\underline{I}n \underline{F}ac\underline{T}orization (\mname) for sparse nonnegative tensors. First we define the input and output for \mname.  Our proposed method requires the $N$-th order tensor $\T{X} \in \mathbb{R}^{I_1 \times ... \times  I_N}$ and $N$ cost matrices $\M{C}_n  \in \mathbb{R}_+^{I_n \times I_n}~(n=1,...,N)$  capturing the relations between  dimensions along each tensor mode as the \textbf{input} to \mname. Here, $\M{C}_n$ is a cost matrix for mode $n$ and can be computed directly from the tensor input, or derived from  external knowledge. It can also be an identity matrix, meaning that the correlation among features are ignored if the cost matrix is not available. \mname is based on CP decomposition and computes $N$ non-negative factor matrices $\M{A}_n \in \mathbb{R}_+^{I_n \times R}~(n =1,...,N)$ as the \textbf{output}. These factor matrices can then be used for downstream tasks, such as clustering and classification. 

\subsection{Wasserstein Distance for Tensors} 
\begin{definition} \label{def:was_mat}
{\bf Wasserstein Matrix Distance:}
Given a cost matrix $\M{C} \in \mathbb{R}_+^{M \times M}$, the Wasserstein distance between two matrices $\M{A}=[\V{a}_1,...,\V{a}_P ] \in \mathbb{R}_+^{M \times P}$ and $ \M{B}=[\V{b}_1,...,\V{b}_P ] \in \mathbb{R}_+^{M \times P}$ is denoted by $W_M ( \M{A},  \M{B} )$, and given by: %
\begin{equation}
\small
\begin{aligned}
 W_M(\M{A}, \M{B}) &= \sum_{p=1}^{P} W_V(\V{a}_p, \V{b}_p) \\
 &= \underset{ \M{T}_p  \in U(\V{a}_p, \V{b}_p)}{\operatorname{minimize}} \sum_{p=1}^{P} \langle \M{C},\M{T}_p \rangle - \frac{1}{\rho} E(\M{T}_p)\\
 &=\underset{ \overline{\M{T}}\in U(\M{A}, \M{B}) }{\operatorname{minimize}}  \langle \overline{\M{C}},\overline{\M{T}} \rangle - \frac{1}{\rho} E(\overline{\M{T}}),
\end{aligned}
\label{Def1}
\end{equation}
Note that sum of the minimization equals minimization of sums since each $\M{T}_p$ is independent of others. Here, $\rho$ is a regularization parameter,  $\overline{\M{C}}= \underbrace{[\M{C},....,\M{C}]}_\text{ $P$ times}$ and $\overline{\M{T}}=[\M{T}_1,...\M{T}_p,...,\M{T}_P]$ are concatenations of the cost matrices and the transport matrices for the $P$ optimal transport problems, respectively. 
Note that 
$U(\M{A}, \M{B})$ is the feasible region of the transport matrix $\overline{\M{T}}$ and is given by:
\begin{equation}\small
\begin{aligned}
    U(\M{A}, \M{B}) &= \left\{ \overline{\M{T}} \in \mathbb{R}_+^{  M \times MP } ~|~ \M{T}_p \V{1}_M =\V{a}_p,  \M{T}^T_p \V{1}_M=\V{b}_p \quad  \forall p \right\} \\
     &= \left\{ \overline{\M{T}} \in \mathbb{R}_+^{  M \times MP } ~|~ \Delta(\overline{\M{T}}) =\M{A}, \Psi(\overline{\M{T}})= \M{B} \right\} 
\end{aligned}
\end{equation}
where $\Delta(\overline{\M{T}})=[\M{T}_1 \V{1}_M,...,\M{T}_P \V{1}_M] = \overline{\M{T}} (\M{I}_P\otimes \boldsymbol{1}_M) $, $\Psi(\overline{\M{T}})= [\M{T}^T_1 \V{1}_M,...,\M{T}^T_P \V{1}_M]$ and $\V{1}_M$ is a one vector with length $M$. 
\label{def:W_M}
\end{definition}

\begin{definition} \label{def:was_ten}
{\bf Wasserstein Tensor Distance:}
The Wasserstein distance between $N$-th order tensor $\T{X} \in \mathbb{R}_+^{I_1 \times ... \times  I_N}$ and its reconstruction $\hat{\T{X}} \in \mathbb{R}_+^{I_1 \times ... \times  I_N}$ is denoted by $W_T(\T{\hat{X}},\T{X})$:%
\begin{equation}\small
\begin{aligned}
W_T(\T{\hat{X}},\T{X}) &= \sum_{n=1}^{N} W_M \Big(  \widehat{\M{X}}_{(n)},\M{X}_{(n)} \Big) \\
&\equiv \sum_{n=1}^N \left\{\underset{ \overline{\M{T}}_n \in U\left( \widehat{\M{X}}_{(n)},\M{X}_{(n)}\right) }{\operatorname{minimize}} ~~ \langle \overline{\M{C}}_n,\overline{\M{T}}_n \rangle - \frac{1}{\rho} E(\overline{\M{T}}_n) \right\},
\label{wass_def}
\end{aligned}
\end{equation}
where %
$\overline{\M{C}}_{n} =[\M{C}_n, \M{C}_n,... ,\M{C}_n ] \in \mathbb{R}_+^{I_n \times I_{n}I_{(-n)}} $ 
is obtained by repeating the cost matrix of the $n$-th mode for $I_{(-n)}$ times and horizontally concatenating them. $\overline{\M{T}}_{n} =[\M{T}_{n1},..., \M{T}_{nj},...,\M{T}_{nI_{(-n)}}] \in \mathbb{R}_+^{I_n \times I_{n}I_{(-n)}} $ and $\M{T}_{nj} \in \mathbb{R}_+^{I_n \times I_n}$ is the transport matrix between the columns $ \widehat{\M{X}}_{(n)}(:,j) \in \mathbb{R}_+^{I_n}$ and $\M{X}_{(n)}(:,j) \in \mathbb{R}_+^{I_n }$.
\label{def:wass_dist_tensor}
\end{definition}

Note that $\overline{\M{C}}$ and $\overline{\M{C}}_n$  are for notation convenience and we do not keep multiple copies of $\M{C}$ and $\M{C}_n$ in implementation. 

\begin{proposition}  \label{prop:wass_dis_proof}
The Wasserstein distance between tensors \T{X} and \T{Y} denoted by $W_T(\T{X},\T{Y})$ is a valid distance and satisfies the metric axioms as follows: 

\begin{enumerate}
    \item Positivity: $W_T(\T{X},\T{Y}) \geq 0$ 
    \item Symmetry: $W_T(\T{X},\T{Y})=W_T( \T{Y},\T{X} )$ 
    \item Triangle Inequality: $ \forall \T{X},\T{Y},\T{Z} \quad W_T(\T{X},\T{Y}) \leq W_T(\T{X},\T{Z}) + W_T(\T{Z},\T{Y})$
\end{enumerate}
\end{proposition}
 We provide the proof in the appendix. 

\subsection{Wasserstein Tensor Factorization}
Given an input tensor $\T{X}$, \mname aims to find  the low-rank approximation $\widehat{\T{X}}$ such that their  Wasserstein distance in \eqref{wass_def} is minimized. Formally, we solve for $\widehat{\T{X}}$ by minimizing $W_T(\widehat{\T{X}},\T{X})$, where $\widehat{\T{X}} = \llbracket \M{A}_1,\dots,\M{A}_N\rrbracket$ is the CP factorization of $\T{X}$. Together with Definitions~\ref{def:W_M} and \ref{def:wass_dist_tensor}, we have the following optimization problem:
\begin{align}
    \underset{ \{ \M{A}_n\geq 0, \overline{\M{T}}_{n} \}_{n=1}^N }{\operatorname{minimize}} & \;
    \sum_{n=1}^{N}  \left(  \langle \overline{\M{C}}_{n}, \overline{\M{T}}_{n}   \rangle - \frac{1}{\rho} E(\overline{\M{T}}_{n}) \right) \label{obj_func_swift_expanded}\\
    \operatorname{subject~to} &\quad\widehat{\T{X}} = \llbracket \M{A}_1, \dots, \M{A}_N \rrbracket \notag\\
    &\quad\overline{\M{T}}_n\in U(\widehat{\M{X}}_{(n)}, \M{X}_{(n)}) ,~ n=1,\dots,N \notag
\end{align}
where the first constraint  enforces a low-rank CP approximation, the second one ensures that the transport matrices are inside the feasible region. We also interested in imposing non-negativity constraint  on the CP factor matrices for both well-definedness of the optimal transport problem and interpretability of the factor matrices. Similar to prior works on vector-based, and matrix-based Wasserstein distance minimization problems~\cite{frogner2015learning, qian2016non}, in order to handle non-probability inputs, we convert the second hard constraint in \eqref{obj_func_swift_expanded} to soft regularizations by Lagrangian method using the generalized KL-divergence. Together with the fact that  $\widehat{\M{X}}_{(n)}=\M{A}_n (\M{A}_{\odot}^{(-n)})^T$ for CP factorization, we convert \eqref{obj_func_swift_expanded} into the following objective function:
\begin{equation}
\begin{aligned}
& \underset{ \{ \M{A}_n\geq 0, \overline{\M{T}}_{n} \}_{n=1}^N  }{\operatorname{minimize}}
 \sum_{n=1}^{N} \Bigg(  \underbrace{\langle \overline{\M{C}}_{n}, \overline{\M{T}}_{n}   \rangle - \frac{1}{\rho} E(\overline{\M{T}}_{n})}_\text{Part $P_1$ } +  \\ 
& \lambda \Big( \underbrace{KL(\Delta(\overline{\M{T}}_{n}) ||  \M{A}_n (\M{A}_{\odot}^{(-n)})^T)  }_\text{Part $P_2$} + \underbrace{ KL(\Psi(\overline{\M{T}}_{n}) || \M{X}_{(n)})}_\text{Part $P_3$}  \Big) \Bigg)  \\
\end{aligned}
\label{SWIFT_obj_lagrangian}
\end{equation}
where $\Delta(\overline{\M{T}}_n) = [\M{T}_{n1} \V{1},...,\M{T}_{nj} \V{1},...,\M{T}_{nI_{(-n)}} \V{1}]$, $\Psi(\overline{\M{T}}_n) = [\M{T}_{n1}^T \V{1},...,\M{T}_{nj}^T \V{1},...,\M{T}_{nI_{(-n)}}^T \V{1}]$, and $\lambda$ is the weighting parameter for generalized KL-divergence regularization.

We use the alternating minimization to solve \eqref{SWIFT_obj_lagrangian}.   \mname iteratively updates: 1) \textbf{Optimal Transports Problems.} For each mode-$n$ matricization, 
\mname computes at most a set of $I_{(-n)}$ optimal transport problems. By  exploiting the sparsity structure and avoiding explicitly computing transport matrices we can significantly reduce the computation cost. 2) \textbf{Factor Matrices.} The CP factor matrices are involved inside the Khatri-Rao product and needs excessive amount of computation.   However, by rearranging the terms involved in \eqref{SWIFT_obj_lagrangian}, we can efficiently update each factor matrix.  %
Next, we provide efficient ways to update optimal transport and factor matrices in more details.

\subsection{\textbf{Solution for Optimal Transport Problems}} \label{OT_sol}

 For mode-$n$,  $\overline{\M{T}}_{n} =[\M{T}_{n1},..., \M{T}_{nj},...,\M{T}_{nI_{(-n)}}] \in \mathbb{R}_+^{I_n \times I_{n}I_{(-n)}} $  includes a set of $I_{(-n)}$ different  optimal transport problems.  The optimal solution of $j$-th optimal transport problem for mode $n$ is $\M{T}_{nj}^* = \text{diag}(\V{u}_j) \M{K}_n \text{diag}(\V{v}_j) $, where $\M{K}_n =e^{(- \rho \M{C}_n-1)} \in \mathbb{R}_+^{I_n \times I_n }, \V{u}_j,  \V{v}_j \in \mathbb{R}_+^{I_n} $. This requires computing $I_{(-n)}$ transport matrices with size $I_n \times I_n$ in~\eqref{SWIFT_obj_lagrangian}, which is extremely time-consuming. To  reduce the amount of computation, \mname proposes the following three strategies:
 
\textbf{1) Never explicitly computing transport matrices ($\overline{\M{T}}^*_{n}$):}  Although we are minimizing \eqref{SWIFT_obj_lagrangian} with respect to $\overline{\M{T}}_{n}$, we never explicitly compute $\overline{\M{T}}^*_{n}$.  In stead of directly computing the optimal transport matrices $\overline{\M{T}}^*_{n}$, we make use of the constraint 
$\M{T}_{nj}^* \V{1} = \operatorname{diag}(\V{u}_j) \M{K}_n \V{v}_j= \V{u}_j* (\M{K}_n \V{v}_j) $ where $*$ denotes element-wise product. As a result, the following proposition effectively updates objective function~\ref{SWIFT_obj_lagrangian}.
\begin{proposition} \label{prop:ot_update}
 $\Delta(\overline{\M{T}}_n)=[\M{T}_{n1} \V{1},...,\M{T}_{nj} \V{1},...,\M{T}_{nI_{(-n)}} \V{1}] = \M{U}_n * ( \M{K}_n \M{V}_n )$ minimizes~\eqref{SWIFT_obj_lagrangian} where $\tiny \M{U}_n= (\widehat{\M{X}}_{(n)})^{\Phi} \oslash \Big( \M{K}_n  \Big(\M{X}_{(n)}  \oslash (\M{K}_n^T \M{U}_n)   \Big)^\Phi  \Big)^\Phi$,    $\tiny \M{V}_n= \Big( \M{X}_{(n)}  \oslash (\M{K}_n^T \M{U}_n)   \Big)^\Phi$, $\Phi=\frac{\lambda \rho}{ \lambda \rho+1}$, and $\oslash$ indicates element-wise division. 
 See Section appendix for proof.
\end{proposition}

\textbf{2) Exploiting Sparsity Structure in $\M{X}_{(n)} \in \mathbb{R}_+^{I_n \times I_{(-n)}}$:} We observe that there are many columns with all zero elements in $\M{X}_{(n)}$ due to the sparsity structure in the input data. There is no need to compute transport matrix for those zero columns, therefore,  we can easily drop zero value columns in $\M{X}_{(n)}$ and its corresponding columns in $\M{U}_n, \M{V}_n,$ and $\widehat{\M{X}}_{(n)}$ from our computations. We use $NNZ_n$ to denote the number of non-zero columns in $\M{X}_{(n)}$. By utilizing this observation, we reduce the number of times to solve the optimal transport problems from $I_{(-n)}$ to $NNZ_n$, where we usually have $NNZ_n \ll I_{(-n)}$ for sparse input.

\textbf{3) Parallelization of the  optimal transport computation:}   The $NNZ_n$  optimal transport problems for each factor matrix $\M{X}_{(n)}$ can be solved independently.  Therefore, parallelization on multiple processes is straightforward for \mname.

\subsection{\textbf{Solution for Factor Matrices}} \label{Fac_mat_sol}

All the factor matrices are  involved in Part $P_2$ of Objective~\eqref{SWIFT_obj_lagrangian} and present in $N$ different KL-divergence terms. The objective function with respect to factor matrix $\M{A}_n$ for mode $n$ is:
\begin{equation}
\begin{aligned}
& \underset{ \M{A}_n \geq 0   }{\operatorname{minimize}} \quad
 \sum_{i=1}^{N}   KL \Big(\Delta(\overline{\M{T}}_i) ~||~ \M{A}_i (\M{A}_{\odot}^{(-i)})^T \Big)
\end{aligned}
\label{obj_factors_simplified}
\end{equation}
where $\Delta(\overline{\M{T}}_i) \in \mathbb{R}_+^{I_i \times I_{(-i)}}$, $\M{A}_i \in \mathbb{R}_+^{I_i \times R}$, $ \M{A}_{\odot}^{(-i)} \in \mathbb{R}_+^{I_{(-i)} \times R} $. Updating factor matrix $\M{A}_n$ in \eqref{obj_factors_simplified} is expensive due to varying positions of $\M{A}_n$ in the $N$ KL-divergence terms. Specifically, $\M{A}_n$ is involved in the Khatri-Rao product $\M{A}_{\odot}^{(-i)}$, as defined in~\eqref{eq:khatri_rao}, for every $i\neq n$. On the other hand, when $i=n$, $\M{A}_n$ is not involved in $\M{A}_{\odot}^{(-i)}$.

\textbf{Efficient rearranging operations.}  In order to solve~\eqref{obj_factors_simplified} efficiently, we introduce operator $\Pi$, which performs a sequence of reshape, permute and another reshape operations, such that, when applied to the right-hand side of~\eqref{obj_factors_simplified}, $\M{A}_n$ is no longer involved inside the Khatri-Rao product for all $i\neq n$. Formally, 
\begin{equation}\small
    \Pi(\M{A}_i (\M{A}_{\odot}^{(-i)})^T,n)=\M{A}_n (\M{A}_{\odot}^{(-n)})^T \in \mathbb{R}_+^{I_n \times I_{(-n)}}~~\forall~i\neq n.
\end{equation}

To maintain equivalence to~\eqref{obj_factors_simplified}, we apply the same operation to the left-hand side of~\eqref{obj_factors_simplified}, which leads us to the following formulation:
\begin{equation}
\small
\begin{aligned}
& \underset{ \M{A}_n \geq 0  }{\operatorname{minimize}}
\quad  KL \Bigg( \begin{bmatrix}
 \Pi(\Delta(\overline{\M{T}}_1),n) \\
. \\
. \\
\Pi(\Delta(\overline{\M{T}}_i),n)\\
. \\
. \\
\Pi(\Delta(\overline{\M{T}}_N),n)
\end{bmatrix}~\Bigg|\Bigg|~ \begin{bmatrix}
 \M{A}_n (\M{A}_{\odot}^{(-n)})^T \\
. \\
. \\
\M{A}_n (\M{A}_{\odot}^{(-n)})^T \\
. \\
. \\
\M{A}_n (\M{A}_{\odot}^{(-n)})^T 
\end{bmatrix}  \Bigg) \\ %
\end{aligned}
\label{An_efficient_format}
\end{equation}

Where $\Pi(\Delta(\overline{\M{T}}_i),n) \in \mathbb{R}_+^{I_n \times I_{(-n)}} $ for all $i$. Due to the fact that KL-divergence is computed point-wisely, the above formula is equivalent to \eqref{obj_factors_simplified}, with the major difference that $\M{A}_n$ is at the same position in every KL-divergence term in~\eqref{An_efficient_format}, and is no longer involved inside the Khatri-Rao product terms; therefore, it can be much more efficiently updated via multiplicative update rules~\cite{lee2001algorithms}. More details regarding operator $\Pi$, and multiplicative update rules for $\M{A}_n$ are provided in the appendix.

In every iteration of \mname, we first update $N$ different optimal transport problems and then update $N$ factor matrices. Algorithm~\ref{WNTF_alg}  summarizes the optimization procedure in \mname. 

\begin{proposition}
\mname is based on Block Coordinate Descent (BCD) algorithm and guarantees convergence to a stationary point. See detailed proofs in the appendix.  
\end{proposition}

Details regarding complexity analysis are provided in appendix. 

\begin{algorithm}
\caption{\mname}\label{WNTF_alg}
\SetAlgoLined
\SetKwInOut{Input}{Input}
\SetKwInOut{Output}{Output}
\Input{$\T{X} \in \mathbb{R}_+^{I_1 \times I_2 \times ... \times I_N},~\M{C}_{n},~NNZ_n ~~ n=1,...,N$, \\~~~~target rank $R$, $\lambda$, and $\rho$}
\Output{$\M{A}_n \in \mathbb{R}_+^{I_n \times R}~~n=1,...,N$}
$\M{K}_n =e^{(- \rho \M{C}_n-1)}~~n=1,...,N$\;
Initialize $\M{A}_n~~n=1,...,N$ randomly\;
\While{stopping criterion is not met}{
    \For{n = 1,...,N}{
        \text{// Optimal Transport Update (Section~\ref{OT_sol})}\;
        $\Phi$ = $\frac{\lambda \rho}{ \lambda \rho+1}$\; 
        $\M{U}_n=ones(I_n, NNZ_n) \oslash I_n$\;
        \For{s=1,...,\textit{Sinkhorn Iteration}}{
            \mbox{$\M{U}_n=(\widehat{\M{X}}_{(n)})^{\Phi} \oslash \Big( \M{K}_n  \Big( \M{X}_{(n)} \oslash (\M{K}_n^T \M{U}_n)   \Big)^\Phi  \Big)^\Phi$\;}
        }
        $\M{V}_n= \Big( \M{X}_{(n)} \oslash (\M{K}_n^T \M{U}_n)   \Big)^\Phi$\;
        $\Delta(\overline{\M{T}}_{n}) = \M{U}_n *(\M{K}_n \M{V}_n)$\;
    }
    \For{$n$=$1,...,N$}{
        \text{// Factor Matrix Update (Section~\ref{Fac_mat_sol})}\;
        Update $\M{A}_n$ based on~\eqref{An_efficient_format};
    }
}
\end{algorithm}

\section{Experimental Results}

\subsection{Experimental Setup}
\subsubsection{Datasets and Evaluation Metrics\\} \label{dataset_descrip}  
\par 1) \textbf{BBC News} \cite{greene06icml} is a publicly available dataset from the BBC News Agency for text classification task.  A third-order count tensor is constructed with the size of 400 articles by 100 words by 100 words. $\T{X}(i,j,k)$ is the number of co-occurrences of the $j$-th and the $k$-th words in every sentence of the $i$-th article. We use the pair-wise cosine distance as the word-by-word and article-by-article cost matrices with details provided in the appendix. The downstream task is to predict the category (from business, entertainment, politics, sport or tech) of each article and we use \textit{accuracy} as the evaluation metric.

\par 2) \textbf{Sutter} is a dataset collected from a large real-world health provider network containing the electronic health records (EHRs) of patients. A third-order binary tensor is constructed with the size of 1000 patients by 100 diagnoses by 100 medications. The downstream task is to predict the onset of heart failure (HF) for the patients (200 out of the 1000 patients are diagnosed with HF) and use PR-AUC (Area Under the Precision-Recall Curve) to evaluate the HF prediction task. 

We chose these two datasets because of different data types (count in BBC, binary in Sutter). Note that  {\bf the cost matrices are derived from the original input by cosine similarity without any additional external knowledge}. Hence the comparisons are fair since the input are the same. %

\subsubsection{Baselines} \label{baselines}

We compare the performance of \mname with different tensor factorization methods with different loss functions and their variants:
\begin{itemize}[leftmargin=*]
    \item The first loss function  minimizes sum square loss  and has 4 variants: 1) \textbf{CP-ALS} \cite{bader2007efficient}; 2) \textbf{CP-NMU} \cite{bader2007efficient}; 3) \textbf{Supervised CP} \cite{kim2017discriminative}; and 4) \textbf{Similarity based CP} \cite{kim2017discriminative}. The first one is unconstrained, the second one incorporates non-negativity constraint, the third one utilizes label information, and the last one uses similarity information among features (similar to \mname). 
\item The second loss function is Gamma loss (\textbf{CP-Continuous} \cite{hong2020generalized}) which is the start of the art method (SOTA) for non-negative continuous tensor.
\item The third loss function is Log-loss (\textbf{CP-Binary} \cite{hong2020generalized}) which is SOTA binary tensor factorization by fitting Bernoulli distribution. 
\item The fourth loss function is Kullback-Leibler Loss (\textbf{CP-APR} \cite{chi2012tensors}) which fits Poisson distribution on the input data and is suitable for count data.
\end{itemize}

\begin{table*}[h]
  \centering
  \caption{\footnotesize The first  part reports  the average and standard deviation of \textit{accuracy}  on the test set as  for different value of R on BBC NEWS data. The second part depicts the average and standard deviation of \textit{PR-AUC Score} on test data for Sutter dataset.  Both experiments are based on five-fold cross validation. We used Lasso Logistic Regression as a classifier.}
    \begin{tabular}{clrrrrr}
          &       & \multicolumn{1}{c}{R=5} & \multicolumn{1}{c}{R=10} & \multicolumn{1}{c}{R=20} & \multicolumn{1}{c}{R=30} & \multicolumn{1}{c}{R=40}  \\
    \toprule
    \multirow{7}[2]{*}{\makecell{\textbf{BBC News} \\ \textbf{Dataset}}} 
          & CP-ALS & .521  $\pm$ .033 &  .571  $\pm$ .072 &  .675 $\pm$ .063 &    .671  $\pm$ .028 &     .671  $\pm$ .040     \\[1pt]
          & CP-NMU &  .484 $\pm$ .039 & .493  $\pm$  .048 &   .581  $\pm$   .064  & .600   $\pm$  .050 &  .650 $\pm$  .031 \\[1pt]
          & Supervised CP & .506  $\pm$ .051 & .625 $\pm$ .073 & .631 $\pm$ .050 & .665 $\pm$ .024 &  .662$\pm$   .012  \\[1pt]
          & Similarity Based CP & .518 $\pm$ .032 & .648 $\pm$ .043 & .638 $\pm$ .021 & .662 $\pm$ .034 & .673 $\pm$ .043  \\[1pt]
          & CP-Continuous & .403 $\pm$ .051  & .481 $\pm$ .056 & .528 $\pm$ .022 & .559 $\pm$ .024 & .543 $\pm$ .043 \\[1pt]
          & CP-Binary &  .746 $\pm$ .058 & .743 $\pm$ .027 & .737  $\pm$ .008 & .756 $\pm$ .062 & .743   $\pm$   .044 \\[1pt]
          & CP-APR & .675 $\pm$ .059 &  .768  $\pm$ .033 &  .753   $\pm$   .035 &   .743 $\pm$ .033 &  .746   $\pm$ .043  \\[1pt]
          & \mname &  \textbf{.759 $\pm$ .013} &  \textbf{.781 $\pm$ .013}  & \textbf{.803 $\pm$ .010} &    \textbf{.815 $\pm$  .005} & \textbf{.818  $\pm$ .022} \\
    \midrule
    \multirow{7}[2]{*}{\makecell{\textbf{Sutter} \\ \textbf{Data}}}  
          & CP-ALS  & .327 $\pm$ .072  &  .333 $\pm$ .064 & .311 $\pm$ .068 & .306 $\pm$ .065 &  .332 $\pm$ .098 \\[1pt]
          & CP-NMU  & .300 $\pm$ .054  & .294 $\pm$ .064 & .325 $\pm$ .085 & .344 $\pm$ .068 & .302 $\pm$ .071 \\[1pt]
          & Supervised CP  & .301 $\pm$ .044  & .305 $\pm$ .036 & .309 $\pm$ .054 & .291 $\pm$ .037 & .293 $\pm$ .051 \\[1pt]
          & Similarity Based CP  & .304 $\pm$ .042  & .315 $\pm$ .041 & .319 $\pm$ .063 & .296 $\pm$ .041 & .303 $\pm$ .032 \\[1pt]
          & CP-Continuous  & .252 $\pm$ .059  & .237 $\pm$ .043 & .263 $\pm$ .065 & .244 $\pm$ .053 & .256 $\pm$ .077 \\[1pt]
          & CP-Binary  & .301 $\pm$ .061  & .325 $\pm$ .079 & .328 $\pm$ .080 & .267 $\pm$ .074 & .296 $\pm$ .063 \\[1pt]
          & CP-APR  & .305 $\pm$ .075  & .301 $\pm$ .068 & .290 $\pm$ .052 & .313 $\pm$ .082 & .304 $\pm$ .086 \\[1pt]
          & \mname & \textbf{.364 $\pm$ .063}  & \textbf{.350 $\pm$ .031} & \textbf{.350 $\pm$ .040 }& \textbf{.369 $\pm$ .066} & \textbf{.374 $\pm$ .044} \\[1pt]

    \bottomrule
    \end{tabular}%
  \label{tab:pred_power_datasets}%
\end{table*}%

\subsection{Classification Performance of \mname} \label{classification_part}
To evaluate low-rank factor matrices, we utilize the downstream prediction tasks as a proxy to assess the performance of \mname and the baselines, similar to the existing works~\cite{ho2014marble,yin2019learning,afshar2020taste,yin2020logpar}.  We performed 5-fold cross validation and split the data into training, validation, and test sets by a ratio of 3:1:1.  %

\par \textbf{Outperforming various tensor factorizations}: Table~\ref{tab:pred_power_datasets} summarizes the classification performance using the factor matrices obtained by \mname and the baselines with varying target rank ($R \in \{5, 10, 20, 30, 40\}$). We report the mean and standard deviation of \textit{accuracy} for BBC News, and that of \textit{PR-AUC Score} for Sutter dataset over the test set. For BBC News, \mname outperforms all baselines for different target ranks with relative improvement ranging from $1.69\%$ to $9.65\%$. For Sutter, \mname significantly outperforms all baselines for all values of $R$ with relative improvement ranging from $5.10\%$ to $11.31\%$. 

\par \textbf{Outperforming various classifiers}: We further compare the performance of \mname against widely-adopted classifiers, including Lasso Logistic Regression, Random Forest, Multi-Layer Perceptron, and K-Nearest Neighbor, using raw data. The input for BBC and Sutter to these classifiers are obtained by matricizing the input tensors along the article mode and the patient mode, respectively.
Table~\ref{tab:performance_on_raw} summarizes the results, and it clearly shows that \mname using Lasso LR classifier even with R=5 outperforms all the other classifiers compared. 

\begin{table}
\centering
\caption{Average and standard deviation of accuracy on BBC NEWS and PR-AUC score  on Sutter data sets by performing Lasso LR, RF, MLP, and KNN  on raw data sets.}
\begin{tabular}{lcc} 
              & \multicolumn{1}{l}{Accuracy on BBC} & \multicolumn{1}{l}{PR-AUC on Sutter} \\\toprule
    Lasso LR  & .728 $\pm$ .013    & .308 $\pm$ .033 \\
    RF & .6281 $\pm$ .049   & .318 $\pm$  .083 \\
    MLP & .690 $\pm$ .052     & .305 $\pm$  .054  \\
    KNN & .5956 $\pm$ .067     & .259 $\pm$ .067 \\\midrule
    \mname  (R=5)            & .759  $\pm$ .013    & .364 $\pm$  .063 \\
    \mname  (R=40)            & \textbf{.818  $\pm$ .020}    & \textbf{.374 $\pm$  .044 }\\\bottomrule
\end{tabular}
\label{tab:performance_on_raw}
\end{table}

\subsection{Classification Performance on Noisy Data}
To measure the performance of \mname against noisy input, we inject noise to the raw input tensor to construct the noisy input tensor. For the binary tensor input, we add Bernoulli noise. Given the noise level ($p$), we randomly choose zero elements, such that the total number of selected zero elements equals to the number of non-zero elements. Then, we flip the selected zero elements to one with probability $p$. We follow the similar procedure for the count tensor input, except that we add the noise by flipping the selected zero value to a count value with probability $p$, and the added noise value is selected uniformly at random between 1 and maximum value in the tensor input. 

\begin{figure}
\centering
\includegraphics[width=0.98\columnwidth]{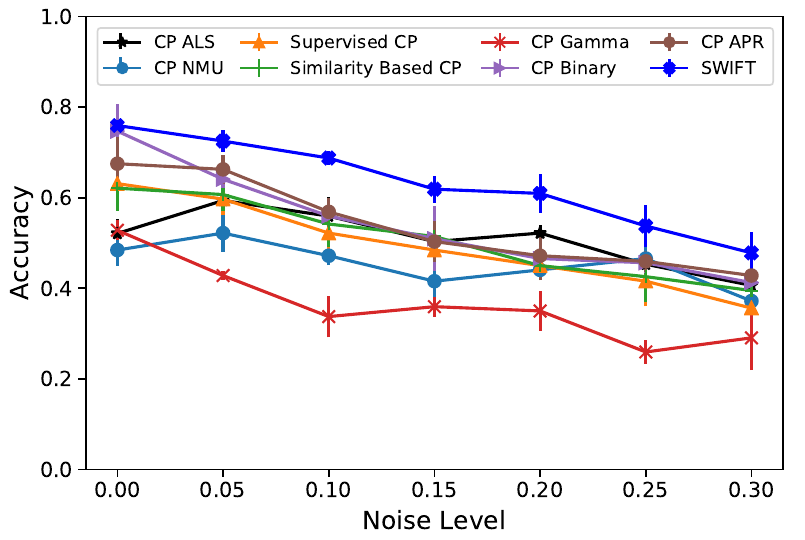}
\caption{\footnotesize The  average and standard deviation of  accuracy  of different baselines as a function of the noise level on BBC NEWS. \mname outperforms other baselines by improving accuracy  up to $15\%$.}\label{fig:noise_BBC}
\end{figure}

\textbf{Performance on BBC data with Noise}:  Figure~\ref{fig:noise_BBC} presents the average and standard deviation of categorizing the articles on the test data with respect to different levels of noise ($p\in \{ 0.05, 0.1,0.15,0.20,0.25,0.30\}$). For all levels of noise, \mname outperforms all baselines by improving accuracy up to $15\%$  over the best baseline especially for medium and high noise levels. Similar results are also achieved on Sutter dataset, as shown in the appendix. 

\subsection{Scalability of \mname} \label{scalability_Sutter}
\begin{figure}
\centering
\begin{subfigure}[]{0.33\textwidth}
    \includegraphics[height=0.4in]{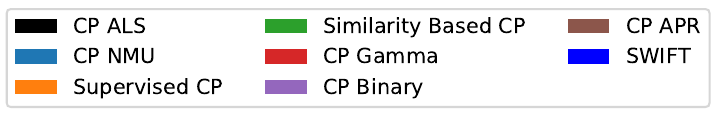}
 \end{subfigure}\\
\begin{subfigure}[]{.4\textwidth}
    \centering
    \includegraphics[height=1.8in]{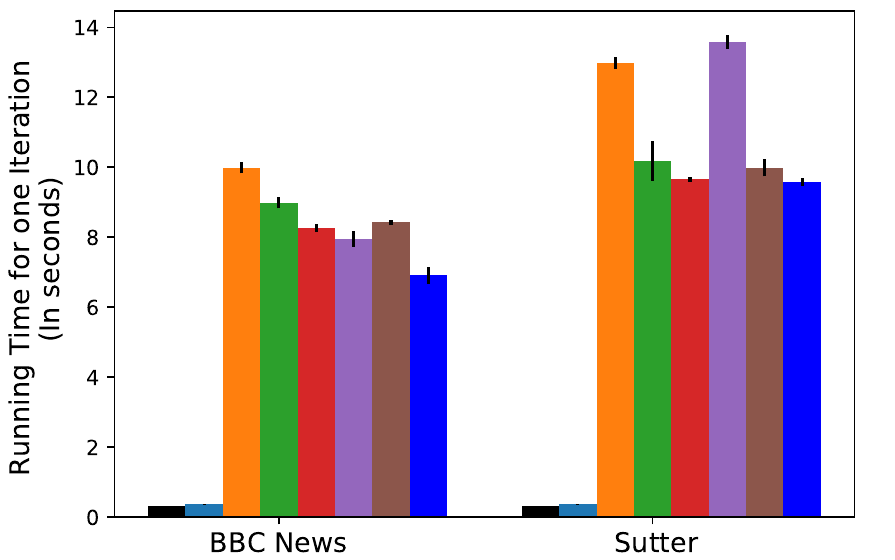}
 \end{subfigure} \\
\caption{The running time of one iteration in seconds on BBC News and Sutter with $R=40$.}\label{Scalability_all}
\end{figure}

In this section, we assess the scalability of \mname in comparison to the other 7 CP algorithms introduced earlier. Figure~\ref{Scalability_all} depicts the average and standard deviation of running time of one iteration in seconds. In this experiment, we set $R=40$. In order to have fair comparisons, we switched off the  parallelization  of  the  optimal  transport computation  for \mname since none of the baselines can be run with parallelization. As shown in Figure~\ref{Scalability_all}, \mname is as scalable as other baselines, suggesting that the great improvement of the performance are achieved without sacrificing the efficiency.  %
We further compare against a direct implementation of Wasserstein tensor factorization. Results are summarized in the appendix, which show that \mname is up to $293x$ faster than the direct implementation. This verifies that the strategies introduced in Section~\ref{swift_framwork} are the keys to \mname being scalable and performant.

\subsection{Interpretability of \mname} \label{interpratability_swift}
To demonstrate the interpretability of results produced by \mname we perform computational phenotyping using Sutter dataset.

\textbf{Computational phenotyping} is a fundamental task in healthcare. It refers to extracting meaningful and interpretable medical concepts (patient clusters) from noisy electronic health records (EHRs)~\cite{richesson2016clinical}. Tensor factorization techniques are powerful tools for extracting meaningful phenotyping~\cite{ho2014marble,yin2018joint,yin2020learning,zhao2019detecting,afshar2020taste,yin2020logpar}.
Here we extract heart failure (HF) phenotypes using Sutter dataset.  We use the same tensor as we described in the previous section and run \mname by selecting $R=40$ since it achieves the highest PR-AUC in comparison to other values of $R$.  $\M{A}_2(:,r), \M{A}_3(:,r)$ represent the membership value of diagnosis and medication features in $r$-th phenotype.

\textbf{Results}: Table~\ref{tab:phen_Sutter} lists three examples of the discovered phenotypes. The weight next to phenotype index indicates the lasso logistic regression coefficient for heart failure prediction (i.e.,21.93, 19.58 and -16.22 in Table~\ref{tab:phen_Sutter}).  ``Dx'' indicates diagnoses and ``Rx'' represents medications.  All phenotypes are clinically meaningful, endorsed and annotated by a medical expert.  The first phenotype is about Atrial Fibrillation which captures patients with  hypertension and cardiac dysrhythmias that are associated high blood pressure medications. Cardiometabolic disease is another phenotype that captures diabetes patients with hypertension.  These two phenotypes have  high positive weights (21.93 and 19.58 respectively) for predicting HF diagnosis. The third phenotype is depressive and menopausal disorders, Serotonin and Benzodiazepines appeared in this phenotype are medications that are commonly prescribed to patients with these conditions in clinical practice.  This phenotype has a negative association with HF (weight = -16.22). The remaining phenotypes positively associated with HF are listed in the appendix.

\begin{table}[]
  \centering
  \caption{ \footnotesize Three phenotypes examples learned on Sutter dataset with \mname. The weight value for each phenotype represents the lasso logistic regression coefficient for the heart failure prediction task. ``Dx'' represents for diagnoses and ``Rx'' indicates for medications. All phenotypes are considered clinically meaningful by a clinical expert.}
    \begin{tabular}{l}
    \toprule
    \textbf{Atrial Fibrillation (Weight= 21.93)} \\
    \midrule
    \textcolor{red}{Dx-Essential hypertension [98.]} \\
    \textcolor{red}{Dx-Disorders of lipid metabolism [53.]} \\
    \textcolor{red}{Dx-Cardiac dysrhythmias [106.]} \\
    \textcolor{blue}{Rx-Calcium Channel Blockers} \\
    \textcolor{blue}{Rx-Alpha-Beta Blockers} \\
    \textcolor{blue}{Rx-Angiotensin II Receptor Antagonists} \\
    \toprule
    \textbf{Cardiometablic Disease (Weight= 19.58)} \\
    \midrule
    \textcolor{red}{Dx-Diabetes mellitus without complication [49.]} \\
    \textcolor{red}{Dx-Essential hypertension [98.]} \\
    \textcolor{red}{Dx-Disorders of lipid metabolism [53.]} \\
    \textcolor{blue}{Rx-Diagnostic Tests} \\
    \textcolor{blue}{Rx-Biguanides} \\
    \textcolor{blue}{Rx-Diabetic Supplies} \\
    \toprule
    \textbf{Mental Disorder (Weight= -16.22)} \\
    \midrule
    \textcolor{red}{Dx-Anxiety disorders [651]} \\
    \textcolor{red}{Dx-Menopausal disorders [173.]} \\
    \textcolor{red}{Dx-Depressive disorders [6572]} \\
       \textcolor{blue}{Rx-Benzodiazepines} \\
    \textcolor{blue}{Rx-Selective Serotonin Reuptake Inhibitors (SSRIs)} \\
    \textcolor{blue}{Rx-Serotonin Modulators} \\
    \bottomrule
    \end{tabular}
  \label{tab:phen_Sutter}
\end{table}

\section{Conclusion}
In this paper, we define the Wasserstein distance between two tensors and propose \mname, an effective tensor factorization model based on the defined tensor Wasserstein distance. To efficiently learn the Wasserstein tensor factorization, we introduce introduce several techniques, including exploitation of the sparsity structure of the input tensor, efficient rearrangement by utilizing tensor properties, and parallelization of the optimal transport computation. Experimental results depict that \mname consistently outperforms baselines in downstream classification tasks for both binary and count tensor inputs. In the presence of noise, \mname also outperforms baselines by a large margin. %

\section*{Acknowledgments}
This work is in part supported by National Science Foundation award SCH-2014438, IIS-1418511, CCF-1533768, IIS-1838042, the National Institute of Health award NIH R01 1R01NS107291-01 and R56HL138415.

\bibstyle{aaai21}
\bibliography{references}

\clearpage
\onecolumn
\appendix

\part{Appendices} %
{\small \parttoc[c]} %

\section{Symbols and Notations} \label{apendx:symbol}
Table~\ref{symbol} presents the symbols and notations used in this paper. 
  \begin{table}[ht] 
      \centering
       \caption{
       Notation and symbols used throughout this paper.
       }
       \label{symbol}
  \begin{tabular}{ c|c}
  \hline
  \hline
  \parbox[t]{1cm}{\centering Symbol} &  \parbox[t]{4cm}{ \centering Definition }  \\
  \hline

  *  & Element-wise multiplication  \\
  $\otimes$ & Kronecker product \\
  $\odot$ &  Khatri–Rao product \\
  $\oslash$ &  Element-wise devision \\
  \hline
  \T{X}, \M{X},\V{x}, x & Tensor, Matrix, vector, scalar \\
  $\M{X}_{(n)}$ & Mode-n Matricization \\
  $\M{A(i,:)}$  & the $i$-th row of $\M{A}$ \\
  $\M{A}(:,r)$ or $\V{a}_r$ & the $r$-th column of $\M{A}$ \\
  $\M{A}_n$ & $n$-th factor matrix \\
  $\M{A}_{\odot}^{(-n)}$ & Khatri-Rao product of all factor matrices except the n-th one \\ 
  $<\V{x}, \V{y}>=\V{x}^T\V{y}$ & Inner product between $\V{x}, \V{y}$ \\
  $\text{vec}(\M{A})$ & converting matrix \M{A} into a vector \\
  \hline
   \hline
 \end{tabular}
 \end{table}

\section{Proofs}
\subsection{Proof of Proposition \ref{prop:wass_dis_proof}}\label{apndx_wass_distance_proof}

For simplicity, we ignore the entropy regularization term ($\frac{1}{\rho} E(\M{T})$) similar to~\cite{cuturi2013sinkhorn}.  In order to prove that Wasserstein distance for tensors is a valid metric, we assume that the cost matrix of mode-$n$ ($\M{C}_{(n)} \in \mathbb{R}_+^{ I_n \times I_n} $):

\begin{itemize}
    \item Is symmetric ($\M{C}_n(i,j)=\M{C}_n(j,i); \quad \forall i,j $).
    \item All of its off diagonal elements are positive and non-zero. ($\M{C}_{(n)}(i,j)=0, \forall i=j$) 
    \item   $\M{C}_{(n)}(i,j) \leq \M{C}_{(n)}(i,k)+\M{C}_n(k,j)\quad\forall i, j, k $
\end{itemize}

The positivity and symmetry of a distance follow then from
\begin{subequations}
\begin{align}
W_T(\T{X},\T{Y}) &= \sum_{n=1}^{N} W_M \Big( \M{X}_{(n)},\M{Y}_{(n)} \Big) \label{equ:9_1} \\
&=  \sum_{n=1}^{N} \sum_{i_n=1}^{I_{(-n)}} W_V(\V{x}_{ni_n}, \V{y}_{ni_n} ) \label{equ:9_2} \\
&=  \sum_{n=1}^{N} \sum_{i_n=1}^{I_{(-n)}} \langle \M{C}_{(n)},\M{T}_{n i_n} \rangle \label{equ:9_3} \\
&=  \sum_{n=1}^{N} \sum_{i_n=1}^{I_{(-n)}}  W_V(\V{y}_{ni_n}, \V{x}_{ni_n}) \label{equ:9_4} \\
&= \sum_{n=1}^{N} W_M \Big( \M{Y}_{(n)}, \M{X}_{(n)} \Big) \label{equ:9_5} \\
&= W_T(\T{Y},\T{X}) \label{equ:9_6}
\end{align}
\end{subequations}

Here, $\V{x}_{ni_n} \in \mathbb{R}_+^{I_n} $,  $\V{y}_{ni_n} \in \mathbb{R}_+^{I_n} $ are probability vectors and $\norm{\V{x}_{ni_n}}_1=\norm{\V{y}_{ni_n}}_1=1$. \eqref{equ:9_1} and \eqref{equ:9_2} are expanded based on Definitions~\ref{def:was_ten}, \ref{def:was_mat}. \eqref{equ:9_3} is written based on the definition of Wasserstein distance for two vectors. \eqref{equ:9_4} is written based on the fact that $\M{C}_n$ is a symmetric matrix $\forall n$. Similarly, \eqref{equ:9_5} and \eqref{equ:9_6} are written based on Definitions~\ref{def:was_mat}, \ref{def:was_ten}. These equations show that Wasserstein distance for two tensors is symmetric.
The proof of positivity is straightforward, since all the off diagonal elements of $\M{C}_{(n)}$ are positive, therefore $\langle \M{C}_{(n)},\M{T}_{n i_n} \rangle > 0 \quad  \forall i_n, n$ which suggests $W_T(\T{X},\T{Y}) > 0$.

To prove the triangle inequality of Wasserstein distances for tensors we know that 
\begin{align}
    W_T(\T{X},\T{Y})= \sum_{n=1}^{N} \sum_{i_n=1}^{I_{(-n)}} W_V(\V{x}_{ni_n}, \V{y}_{ni_n} )
\end{align}
\begin{align}
    W_T(\T{X},\T{Z})= \sum_{n=1}^{N} \sum_{i_n=1}^{I_{(-n)}} W_V(\V{x}_{ni_n}, \V{z}_{ni_n} )
\end{align}
\begin{align}
    W_T(\T{Z},\T{Y})= \sum_{n=1}^{N} \sum_{i_n=1}^{I_{(-n)}} W_V(\V{z}_{ni_n}, \V{y}_{ni_n} )
\end{align}

We already know that triangle inequality of Wasserstein distances for every three probability vectors holds \cite{cuturi2013sinkhorn, peyre2019computational} which means:
\begin{align}
    \forall  \V{x}_{ni_n}, \V{y}_{ni_n}, \V{z}_{ni_n} \quad W_V(\V{x}_{ni_n}, \V{y}_{ni_n} ) \leq  W_V(\V{x}_{ni_n}, \V{z}_{ni_n} )+  W_V(\V{z}_{ni_n}, \V{y}_{ni_n} )
\end{align}

By knowing this fact we can extend it to every triple of tensors ($\T{X},\T{Y}, \T{Z}$) where their unfolding includes probability vectors: 
\begin{align}
    W_T(\T{X},\T{Y}) \leq W_T(\T{X},\T{Z}) + W_T(\T{Z},\T{Y})
\end{align}

\subsection{Proof of Proposition \ref{prop:ot_update}} \label{proof_prop_1}
Here, we follow the same strategy as \cite{frogner2015learning}. $\M{T}_{nj}$ is the optimal transport matrix between  $\widehat{\M{X}}_{(n)}(:,j), \M{X}_{(n)}(:,j)$  in mode-n.  In order to compute the $\M{T}_{nj}$, we need to take the derivative with respect to \eqref{SWIFT_obj_lagrangian}. 

\begin{equation*}
\begin{aligned}
\frac{\partial  W_T(\T{X},\widehat{\T{X}})}{\partial \M{T}_{nj} (i,k)}= \frac{\partial \Big(  \langle \overline{\M{C}}_{n}, \overline{\M{T}}_{n}   \rangle - \frac{1}{\rho} E(\overline{\M{T}}_{n}) +\lambda \Big(KL( \Delta(\overline{\M{T}}_n) || \widehat{\M{X}}_{(n)}) + KL( \Psi(\overline{\M{T}}_n) || \M{X_{(n)}})  \Big) \Big) }{\partial \M{T}_{nj} (i,k)}&=0 \\
\end{aligned}
\end{equation*}

\begin{equation*}
\begin{aligned}
&\Rightarrow \M{C}_n(i,k)+\frac{1}{\rho} (\text{log}(\M{T}_{nj} (i,k))+1)   + \lambda \text{log} \Big( \M{T}_{nj}\V{1} \oslash \widehat{\M{X}}_{(n)}(:,j) \Big)_i  + \lambda \text{log}  \Big( \M{T}_{nj}^T\V{1} \oslash \M{X}_{(n)}(:,j) \Big)_k=0 \\
&\Rightarrow \text{log}(\M{T}_{nj}(i,k))+ \lambda \rho \bigg(\text{log} \Big( \M{T}_{nj}\V{1} \oslash \widehat{\M{X}}_{(n)})(:,j) \Big)_i  +  \text{log}  \Big( \M{T}_{nj}^T\V{1} \oslash \M{X}_{(n)}(:,j) \Big)_k \bigg) = -\rho \M{C}_n(i,k)-1 \\
&\Rightarrow \M{T}_{nj}(i,k) *  \bigg(\Big( \M{T}_{nj}\V{1} \oslash \widehat{\M{X}}_{(n)}(:,j) \Big)^{\lambda \rho}_i \Big( \M{T}_{nj}^T\V{1} \oslash \M{X}_{(n)}(:,j) \Big)^{\lambda \rho}_k \bigg) = \text{exp}(-\rho \M{C}_n(i,k)-1) \\
&\Rightarrow \M{T}_{nj}(i,k)=  \Big( \widehat{\M{X}}_{(n)}(:,j) \oslash \M{T}_{nj}\V{1}  \Big)^{\lambda \rho}_i   \Big( \M{X}_{(n)}(:,j) \oslash \M{T}_{nj}^T\V{1}  \Big)^{\lambda \rho}_k \text{exp}(-\rho \M{C}_n(i,k)-1) \\
\end{aligned}
\end{equation*}

Therefore, $\M{T}_{nj} = \text{diag}(\V{u}_j)\M{K}_n\text{diag}(\V{v}_j)$ where $\V{u}_j=( \widehat{\M{X}}_{(n)}(:,j) \oslash \M{T}_{nj}\V{1}  )^{\lambda \rho} \in \mathbb{R}^{I_n }  $ and $\V{v}_j=( \M{X}_{(n)}(:,j) \oslash \M{T}_{nj}^T\V{1})^{\lambda \rho} \in \mathbb{R}^{I_n } $ and $\M{K}_n =e^{(- \rho \M{C}_n-1)} $ where $\oslash$ represents element-wise division. 

\begin{equation*}
\begin{aligned}
&\quad\quad\M{T}_{nj}\V{1} = \text{diag}(\V{u}_j)\M{K}_n\V{v}_j  = \V{u}_j*(\M{K}_n\V{v}_j)  = ( \widehat{\M{X}}_{(n)}(:,j) \oslash \M{T}_{nj}\V{1}  )^{\lambda \rho}  *(\M{K}_n\V{v}_j) \\
&\Rightarrow (\M{T}_{nj}\V{1})^{(\lambda \rho+1)} = ( \widehat{\M{X}}_{(n)}(:,j) )^{\lambda \rho}*(\M{K}_n\V{v}_j)  \\
&\Rightarrow \text{diag}(\V{u}_j)(\M{K}_n\V{v}_j)^{(\lambda \rho+1)} = \text{diag}(\widehat{\M{X}}_{(n)}(:,j))^{\lambda \rho}(\M{K}_n\V{v}_j) \\
&\Rightarrow \V{u}_j^{(\lambda \rho+1)} = (\widehat{\M{X}}_{(n)}(:,j))^{\lambda \rho}*(\M{K}_n\V{v}_j)^{(-\lambda
\rho)} \\
&\Rightarrow \V{u}_j = (\widehat{\M{X}}_{(n)}(:,j))^\frac{\lambda \rho}{\lambda \rho+1}*(\M{K}_n\V{v}_j)^\frac{-\lambda
\rho}{(\lambda \rho+1)} \\
&\Rightarrow  \M{U}_n =[\V{u}_1,...,\V{u}_j,...,\V{u}_{I_{(-n)}}] = (\widehat{\M{X}}_{(n)})^\frac{\lambda \rho}{\lambda \rho+1}
*(\M{K}_n\M{V}_n)^\frac{-\lambda
\rho}{(\lambda \rho+1)}  =(\widehat{\M{X}}_{(n)})^\frac{\lambda \rho}{\lambda \rho+1}
\oslash(\M{K}_n\M{V}_n)^\frac{\lambda
\rho}{(\lambda \rho+1)}  
\end{aligned}
\end{equation*}

By applying the similar procedure $\M{V}_n=(\M{X}_{(n)})^\frac{\lambda \rho}{\lambda \rho+1} \oslash (\M{K}_n^T\M{U}_n)^\frac{\lambda
\rho}{(\lambda \rho+1)} $. 
Therefore, $\tiny \M{U}_n= (\widehat{\M{X}}_{(n)})^{\Phi} \oslash \Big( \M{K}_n  \Big(\M{X}_{(n)}  \oslash (\M{K}_n^T \M{U}_n)   \Big)^\Phi  \Big)^\Phi$ where $\Phi=\frac{\lambda \rho}{\lambda \rho+1}$. \\
By knowing $\M{T}_{nj}\V{1}= \V{u}_j*(\M{K}_n\V{v}_j)$ we can extend $\M{\Delta}_n=[\M{T}_{n1} \V{1},...,\M{T}_{nj} \V{1},...,\M{T}_{nI_{(-n)}} \V{1}] = \M{U}_n *(\M{K}_n \M{V}_n)$.

\subsection{Proof of Proposition 3 } \label{apendix_proof_conv}

Consider \mname's objective function:
\begin{equation}
\begin{aligned}
& \underset{ \{ \M{A}_n\geq 0, \overline{\M{T}}_{n} \}_{n=1}^N  }{\operatorname{minimize}}
 \sum_{n=1}^{N} \Bigg(  \underbrace{\langle \overline{\M{C}}_{n}, \overline{\M{T}}_{n}   \rangle - \frac{1}{\rho} E(\overline{\M{T}}_{n})}_\text{Part $P_1$ } +  \lambda \Big( \underbrace{KL(\Delta(\overline{\M{T}}_{n}) ||  \M{A}_n (\M{A}_{\odot}^{(-n)})^T)  }_\text{Part $P_2$} + \underbrace{ KL(\Psi(\overline{\M{T}}_{n}) || \M{X}_{(n)})}_\text{Part $P_3$}  \Big) \Bigg)  \\
\end{aligned}
\label{SWIFT_obj_lagrangian_appendix}
\end{equation}

The variables in \eqref{SWIFT_obj_lagrangian_appendix} separated into $N$ different optimal transport problems ($ \overline{\M{T}}_{n}$) and $N$ different factor matrices ($\M{A}_n$). %
We use  Block Coordinate Descent (BCD) framework. BCD iteratively  solves one variable at a time while fixing others. Therefore, in every step the objective function is minimized with respect to each of the variables~\cite{bertsekas1997nonlinear}. Each sub-problem in \eqref{SWIFT_obj_lagrangian_appendix} is continuous, differentiable, and  strictly convex since minimizing the Frobenius inner product, negative entropy, and KL-Divergence are strictly convex  problems~\cite{boyd2004convex}. Therefore, based on~\cite{bertsekas1997nonlinear}, \eqref{SWIFT_obj_lagrangian_appendix} converges to a stationary point.

\section{Other Technical Details}
\subsection{Details on Updating Factor Matrix $\M{A}_n$ } \label{appendx:update:factor}
Algorithm~\ref{PHI_alg} presents the pseudo code for updating factor matrix $\M{A}_n$. As we mentioned earlier,  operator $\Pi(\Delta(\overline{\M{T}}_i),n)\sim \mathbb{R}^{I_i\times I_{(-i)}}\rightarrow \mathbb{R}^{I_n\times I_{(-n)}} $ executes a sequence of reshape, permute, reshape on $\Delta(\overline{\M{T}}_i) \in \mathbb{R}^{I_i \times I_{(-i)}}$  and converts its size to $I_n \times I_{(-n)} $.  More specifically, line 2 in Algorithm~\ref{PHI_alg} reshapes matrix $\Delta(\overline{\M{T}}_i) \in \mathbb{R}^{I_i \times I_{(-i)}}$ to tensor $\T{D}_i  \in \mathbb{R}^{I_1 \times...\times I_{i-1} \times I_{(i+1)} \times ...I_N \times I_{i}} $. Line 3-9 permute modes $i$, $n$ in $\T{D}_i$ and map it to $\T{D}_{i \rightarrow n} \in \mathbb{R}^{I_1 \times...\times I_{n-1} \times I_{n+1} \times ...I_N \times I_{n}}$. Depending on the position of $i$, Lines 3,4 show the permutation when $i < n$, Lines 5,6 show the case when $i$ equals $n$, and lines 7,8 present the permutation when $i > n$. Here, $idx(.)$ returns the index of a given mode. For instance, in $[I_1,...,I_{i-1},I_{i+1},...,I_N,I_i]$, $\text{idx}(i-1)=i-1, \text{idx}(i)=N,  \text{idx}(i+1)=i$.
In line 10, tensor $\T{D}_{i \rightarrow n}$ reshapes to a matrix with size $I_n \times I_{(-n)}$. 
Once we apply $\Pi$ on both right and left hand side of \eqref{obj_factors_simplified}, we get the new form as follows: 
\begin{align}\label{fomulation:factor:mat}
    \underset{\M{A}_n \geq 0}{\operatorname{minimize}}~ \sum_{i=1}^N KL\left(\Pi(\Delta(\overline{\M{T}}_i), n) ~||~ \M{A}_n(\M{A}_{\odot}^{(-n)})^T \right)
\end{align}

Now we can easily update $\M{A}_n$ based on a multiplicative update rule~\cite{lee2001algorithms} with the following form (line 13 in Algorithm~\ref{PHI_alg}):
\begin{align} \label{fomulation:update_factor}
\M{A}_n = \M{A}_n * \Big( \Big( \Big( \sum_{i=1}^{N} \Pi(\Delta(\overline{\M{T}}_i), n)  \oslash \M{A}_n(\M{A}_{\odot}^{(-n)})^T \Big) \M{A}_{\odot}^{(-n)} \Big) \oslash \M{1} \M{A}_{\odot}^{(-n)} \Big) 
\end{align}

\begin{algorithm}

\caption{ Updating Factor Matrix $\M{A}_n$}\label{PHI_alg}
\SetAlgoLined
\SetKwInOut{Input}{Input}
\SetKwInOut{Output}{Output}

\Input{$\Delta(\overline{\M{T}}_i) \in \mathbb{R}^{I_i \times I_{(-i)}}  \quad \forall  $  $i=1,...,N$, mode-$n$,  $\M{P}=\M{A}_n (\M{A}_{\odot}^{(-n)})^T$}
\Output{$\M{A}_n \in \mathbb{R}^{I_n \times R}$}

\For{i=1,...,N}{

\tcp{Reshape a matrix to a tensor.}
$\T{D}_i=\text{reshape}(\Delta(\overline{\M{T}}_i), [I_1,...,I_{(i-1)},I_{(i+1)},...,I_N,I_i])$ \;

\tcp{Interchange the $i$-th and $n$-th modes of the tensor.}
\uIf {$(i < n)$}{
   $\T{D}_{i \rightarrow n} = \text{Permute}(\T{D}_i, [\text{idx}(1),..,\text{idx}(i-1),\text{idx}(i), \text{idx}(i+1),..,\text{idx}(n-1),\text{idx}(n+1),..,\text{idx}(N), \text{idx}(n)])$  \;}
\uElseIf {$(i == n)$}{
  $\T{D}_{i \rightarrow n}= \T{D}_i$ \;
  }
\Else{
    $\T{D}_{i \rightarrow n} = \text{Permute}(\T{D}_i, [\text{idx}(1),..,\text{idx}(n-1), \text{idx}(n+1),..,\text{idx}(i-1),\text{idx}(i),\text{idx}(i+1),..,\text{idx}(N) ,\text{idx}(n)])$ \;
    }
\tcp{Reshape a tensor to a matrix.}
$\Pi(\Delta(\overline{\M{T}}_i),n)=\text{reshape}(\T{D}_{i \rightarrow n}, [I_n,I_{(-n)}])$\;
$\Pi(\M{A}_i (\M{A}_{\odot}^{(-i)})^T,n)=\M{P} $\;
}

$\M{A}_n = \M{A}_n * \Big( \Big( \Big( \sum_{i=1}^{N} \Pi(\Delta(\overline{\M{T}}_i), n)  \oslash \M{P} \Big) \M{A}_{\odot}^{(-n)} \Big) \oslash \M{1} \M{A}_{\odot}^{(-n)} \Big) $\;
\end{algorithm}

\subsection{Cost Matrix Calculation} \label{COST_MAT_CALC}
The cost matrices are derived from the same input data without any additional external knowledge.

For BBC NEWS data we compute the following cost matrices:
\begin{itemize}[leftmargin=*]
    \item \textbf{Article $\times$ Article:}  The articles  are converted to a matrix of TF-IDF (term frequency-inverse document frequency)  features. Then the cost matrix between articles $i,j$ is computed based on cosine distance ($\M{C}(i,j)=1-\frac{<\V{a}_i,\V{a}_j>}{||\V{a}_i||||\V{a}_j||}$)   where $\V{a}_i$ is the TF-IDF vector of article i . 
    \item \textbf{Word $\times$ Word:} For each word, we construct a multi-hot encoding vector with the size of  documents in the training set. Then we define the cost matrix as the cosine distance between every pair of vectors. 
 \end{itemize}   

The cost matrices in  Sutter data are:

\begin{itemize}[leftmargin=*]
\item \textbf{Patient $\times$ Patient:} For each patient we  create a vector  by concatenating  diagnosis and medication features  and use cosine distance to compute the cost matrix. 
\item \textbf{Diagnosis $\times$ Diagnosis:} We represent each diagnosis  as a multi-hot encoding vector with the size of  patients in the training set. If a patient have a certain diagnosis then its corresponding value is one, zero otherwise. We use cosine distance  to calculate the cost matrix between every pair of diagnosis vectors.  
\item \textbf{Medication $\times$ Medication:}  We perform a similar computation as the Diagnosis $\times$ Diagnosis cost matrix. 

Note that for \textbf{Similarity based CP} \cite{kim2017discriminative}, we use the same cost matrices.
\end{itemize}

\subsection{Complexity Analysis of \mname } \label{apndx_complexity}
Here, we provide the complexity analysis of different parts of \mname.  In every iteration, we update $N$ different optimal transport problems and $N$ factor matrices based on Algorithm~\ref{WNTF_alg}.

In order to update \textbf{$n$-th optimal transport problem ($\overline{\M{T}}_{n}$)}, we need to compute the following parts: 
$\M{K}_n$ requires $\mathcal{O}(I_n^2)$ flops (\textit{i.e.} floating-point operation). Computing  $\M{A}_{\odot}^{(-n)}$  needs $\mathcal{O}( I_{(-n)} R ) $ and $\widehat{\M{X}}_{(n)}=\M{A}_n (\M{A}_{\odot}^{(-n)})^T$ requires $\mathcal{O}( I_1 ... I_N R )$  flops.  The complexity of computing $\M{U}_n= (\widehat{\M{X}}_{(n)})^{\Phi} \oslash \Big( \M{K}_n  \Big( \M{X}_{(n)} \oslash (\M{K}_n^T \M{U}_n)   \Big)^\Phi  \Big)^\Phi$ (line 8 in Algorithm~\ref{WNTF_alg})  is $ \mathcal{O}( I_n^2 NNZ_n) $ where  $ NNZ_n$ indicates the number of non-zero columns in $\M{X}_{(n)}$. Line 10 in Algorithm~\ref{WNTF_alg} ($ \M{V}_n= \Big( \M{X}_{(n)} \oslash (\M{K}_n^T \M{U}_n)   \Big)^\Phi $) requires $\mathcal{O}( I_n^2 NNZ_n ) $ flops and finally computing $\Delta(\overline{\M{T}}_{n}) = \M{U}_n *(\M{K}_n \M{V}_n)$ involves $\mathcal{O}(I_n^2 NNZ_n )$. 

The steps for updating \textbf{Factor matrix $n$} are as follows: Both reshape and permute operations can be done in $ \mathcal{O}(N) $.  Factor matrix $\M{A}_n$ is updated based on \eqref{fomulation:update_factor} which can be computed in $ \mathcal{O}( R I_1...I_N ) $ flops.
Finally, the total complexity of \mname is $\mathcal{O}(R I_1...I_N)$.

\subsection{Illustration of the second and third strategies}
\begin{figure}[h!]
    \centering\includegraphics[width=0.85\textwidth]{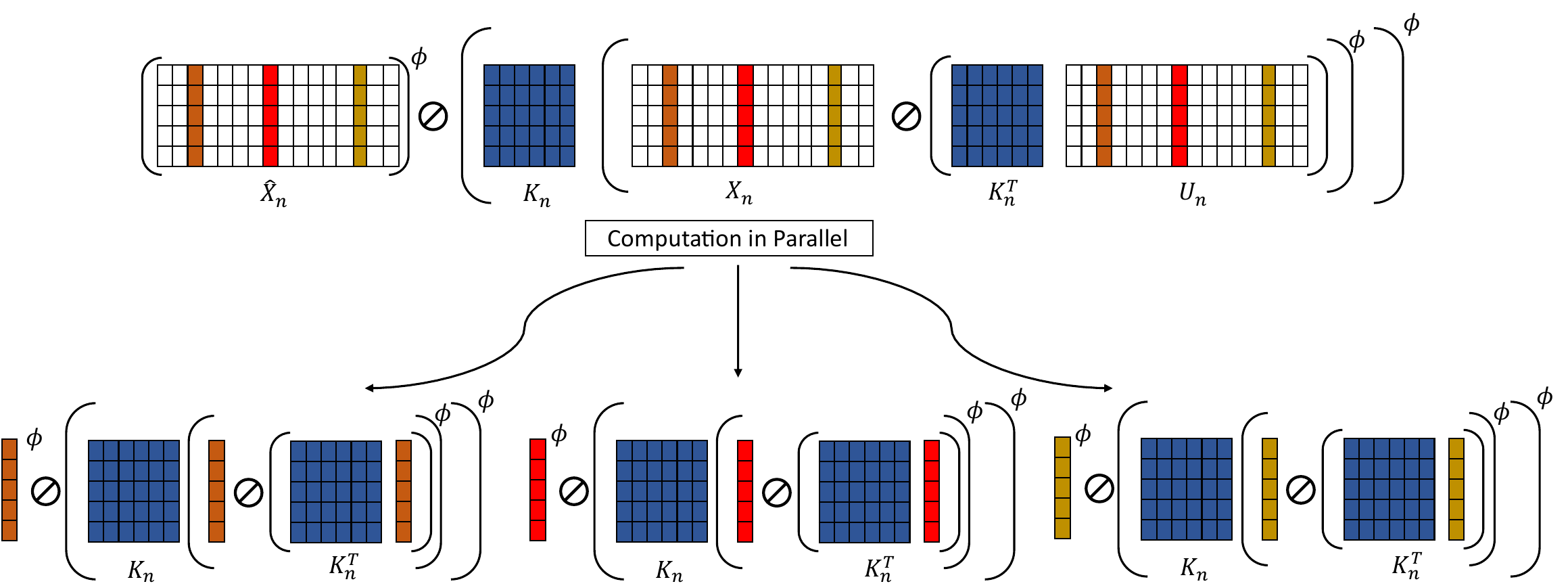}
    \caption{\mname explores sparsity structure in input data $\M{X}_{(n)}$ and drop zero values columns. Also \mname parallelize optimal transport problems for each factor matrix $\M{A}_n$. $NNZ_n$ in this toy example equals 3.}\label{fig:spars_parallel}
\end{figure}

Figure~\ref{fig:spars_parallel} depicts in detail how the second and third strategies introduced in Section~\eqref{OT_sol} explore and utilize the sparsity structure in input tensor in more details.

\section{Additional Experiments and Results}
\subsection{Classification Performance of  \mname. } \label{Q1_appendix}

\par \textbf{Projection on Learned Factor Matrices}
Given the learned factor matrices ($\M{A}_{train_2},...,\M{A}_{train_N}$) from training data, \mname is able to project the new unseen data $\T{X}_{new} \in \mathbb{R}^{I_{new_1} \times I_2 \times ... \times I_N}$ into existing factors and learn $\M{A}_{new_1}$ . 
\begin{equation}
\label{obj_func_swift_projected}
\begin{aligned}
& \underset{ \{ \overline{\M{T}}_{n}\}_{n=1}^N, \M{A}_{new_1} \geq 0  }{\text{minimize}}
& &  \sum_{n=1}^{N}  \left(  \langle \overline{\M{C}}_{n}, \overline{\M{T}}_{n}   \rangle - \frac{1}{\rho} E(\overline{\M{T}}_{n}) \right) \\
& \text{Subject to}
&  & 	\overline{\M{T}}_n\in U(\M{X}_{new(n)}, \widehat{\M{X}}_{new(n)})  \quad \forall n=1,\dots,N, \\
&&& \widehat{\T{X}} = \llbracket \M{A}_{new_1},\M{A}_{train_2} \dots, \M{A}_{train_N} \rrbracket, \\
\end{aligned}
\end{equation}

Here, we need to minimize~\eqref{obj_func_swift_projected} with respect to $\overline{ \M{T}}_{n}$, for $n=1,...,N$ and $\M{A}_{new_1}$ while $\M{A}_{train_2},...,\M{A}_{train_N}$ are fixed. After learning $\M{A}_{new_1}$, we pass it to Lasso Logistic Regression to predict the labels. We use \eqref{obj_func_swift_projected} for projecting the  validation and test data.

\par \textbf{Training Strategy:} Assume we want to classify the elements in the first factor matrix ($\M{A}_1$) (e.g. article mode for BBC News and patient mode for Sutter). For all the approaches under the comparison, we use a similar training strategy with the following steps: 1) we split tensor \T{X} from it's first mode  into training, validation, and test sets by a ratio of 3:1:1 and construct tensors ($\T{X}_{train}$), ($\T{X}_{val}$), and ($\T{X}_{test}$).  2) We train the factorization model using the training set ($\T{X}_{train}$) and compute $\M{A}_{train_n} \quad n=1,...,N$. Note that in this step, the label information is not used (except for the baseline Supervised CP, which use label information in the factorization step). 3) Then we freeze the factor matrices of all modes except the one with label information (article mode for BBC News and patient mode for Sutter), and project the validation and test sets onto the learned factor matrices to obtain the factor matrix of the mode with labels for the test set based on Equation\eqref{obj_func_swift_projected}. 4) Finally we use a lasso logistic regression to perform the classification. We used a five-fold cross validation. 
\par \textbf{\mname Stopping Criteria}
In order to provide scalable solutions, \mname never directly computes transport matrices, therefore, it is expensive to compute the value of its objective function.  Similar to \cite{frogner2015learning,cuturi2013sinkhorn,qian2016non},  we set the number of fixed-point iterations
to an arbitrary number.  Stopping criterion is set to 50 iterations and Sinkhorn iteration to 25.

\par \textbf{Setting hyper-parameters:} For \mname, we execute a grid search for $\lambda \in \{ 0.1,1,10 \}$ and $\rho \in \{ 10,20,50,100,1000 \}$ and regularization parameter for lasso logistic regression ($\eta \in \{ 1e-2, 1e-1, 1, 10, 100, 1000, 10000 \} $). The hyper-parameters of other  baselines including parameter $\eta$ are carefully tuned for each dataset.

\subsection{Performance on Noisy Sutter Data} \label{apndx_perform_noise}

\par
\begin{figure}[h!]
\centering
\begin{subfigure}[H]{0.9\textwidth}\centering
    \includegraphics[width=0.6\textwidth]{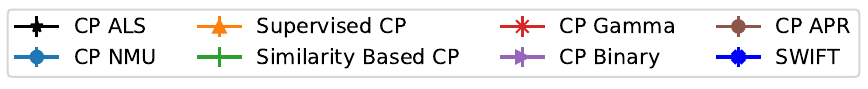}
 \end{subfigure}\\
\begin{subfigure}[H]{.8\textwidth}
    \centering
    \includegraphics[width=0.6\textwidth]{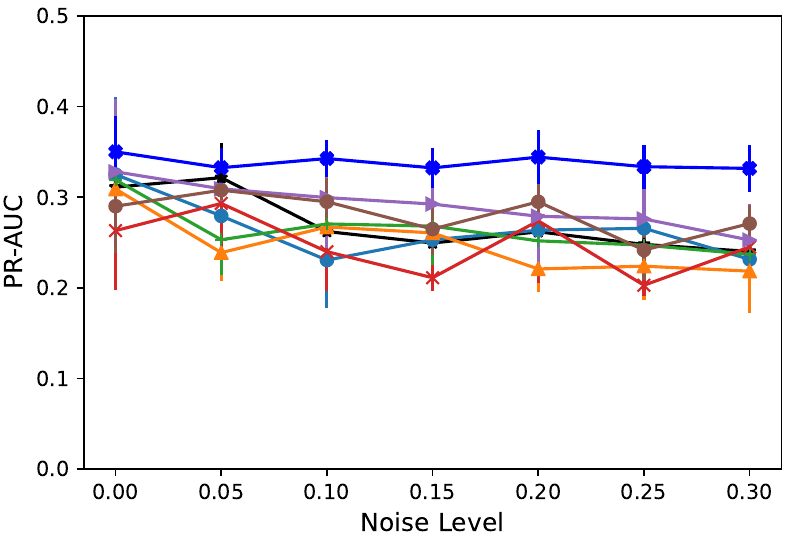}
\end{subfigure}
\caption{\footnotesize The  average and standard deviation of  PR-AUC score of different baselines as a function of the noise level on Sutter data set.} %
\label{fig:noise_Sutter}
\end{figure}

 \textbf{Classification Performance on Noisy Input:}  Figure~\ref{fig:noise_Sutter} presents the average and standard deviation of PR-AUC Score of heart failure prediction on test data in Sutter data set for five-fold cross validation with respect to different levels of noise ($\{ 0.05, 0.1,0.15,0.20,0.25,0.30\}$).    \mname performs better under various noise  levels and relatively improves the  PR-AUC score over
the best baseline by up to $17\%$.   

\subsection{Scalability of \mname on BBC NEWS and Sutter Data} \label{appendix_scalability_swift}
 In this section, we answer the following question: \textbf{Q:} How scalable is \mname more compared to a direct implementation?
\par \textbf{Baseline Construction:} We implement a direct version of Wasserstien tensor factorization for 3-order tensors where for optimal transport computation we do not leverage sparsity structure. For updating factor matrix $\M{A}_n$ in the direct version of Wasserstien tensor factorization we move $\M{A}_n$ into a similar position by vectorizing every $\Delta(\M{T}_i)$ and $\M{A}_n$  and rearranged  the terms using the properties of Khatri-Rao and Kronecker products, so that $\M{A}_n$ has the same position in all $N$ different KL-divergence terms.  We directly utilize Equations \eqref{A1_long_format},\eqref{A2_long_format}, \eqref{A3_long_format} to update $\M{A}_1$, $\M{A}_2$, and $\M{A}_3$, respectively.  Note that direct Wasserstein TF  returns the same solution given the same initialization as to \mname.

Each of these objective functions are equivalent to \eqref{obj_factors_simplified}, \eqref{An_efficient_format} but requires more computation.

\begin{equation}
\small
\begin{aligned}
& \underset{\M{A}_1 \ge 0 }{\text{minimize}}
& & KL \Bigg( \begin{bmatrix}
\text{vec}(\M{\Delta({\M{T}_1})}) \\
\text{vec}(\M{\Delta({\M{T}_2})}^T)  \\
\text{vec}(\M{\Delta({\M{T}_3})}^T)
\end{bmatrix} \Bigg|\Bigg|  \begin{bmatrix}
(\M{A}_3 \odot \M{A}_2) \otimes \M{I}_{I_1} \\
(\M{A}_2 \otimes \M{I}_{I_3I_1})  ((\M{I}_R \odot \M{A}_3) \otimes \M{I}_{I_1}) \\
(\M{A}_3 \otimes \M{I}_{I_2I_1})  ((\M{I}_R \odot \M{A}_2) \otimes \M{I}_{I_1}) \Big)
\end{bmatrix} \text{vec}(\M{A}_1) \Bigg) \\
\end{aligned}
\label{A1_long_format}
\end{equation}

\begin{equation}
\small
\begin{aligned}
& \underset{\M{A}_2 \ge 0 }{\text{minimize}}
& & KL \Bigg( \begin{bmatrix}
\text{vec}(\M{\Delta({\M{T}_1})}^T) \\
\text{vec}(\M{\Delta({\M{T}_2})})  \\
\text{vec}(\M{\Delta({\M{T}_3})}^T)
\end{bmatrix} \Bigg|\Bigg|  \begin{bmatrix}
(\M{A}_1 \otimes \M{I}_{I_2I_3})  ((\M{I}_R \odot \M{A}_3) \otimes \M{I}_{I_2}) \\
(\M{A}_3 \odot \M{A}_{1}) \otimes \M{I}_{I_2}\\
 (\M{A}_3 \otimes \M{I}_{I_1I_2})    \Big( \M{I}_{I_2R} \odot (\M{A}_1(\M{I}_R \otimes \V{1}_{1 \times I_2})) \Big)
\end{bmatrix} \text{vec}(\M{A}_2) \Bigg) \\
\end{aligned}
\label{A2_long_format}
\end{equation}

\begin{equation}
\small
\begin{aligned}
& \underset{\M{A}_3 \ge 0}{\text{minimize}}
& & KL \Bigg( \begin{bmatrix}
\text{vec}(\M{\Delta({\M{T}_1})}^T) \\
\text{vec}(\M{\Delta({\M{T}_2})}^T)  \\
\text{vec}(\M{\Delta({\M{T}_3})})
\end{bmatrix} \Bigg|\Bigg|  \begin{bmatrix}
(\M{A}_1 \otimes \M{I}_{I_3I_2})    \Big( \M{I}_{I_3R} \odot (\M{A}_2(\M{I}_R \otimes \V{1}_{1 \times I_3})) \Big) \\
(\M{A}_2 \otimes \M{I}_{I_3I_1})    \Big( \M{I}_{I_3R} \odot (\M{A}_1(\M{I}_R \otimes \V{1}_{1 \times I_2})) \Big) \\
(\M{A}_2 \odot \M{A}_1) \otimes \M{I}_{I_3}
\end{bmatrix} \text{vec}(\M{A}_3) \Bigg) \\
\end{aligned}
\label{A3_long_format}
\end{equation}

 We can easily prove that equations \eqref{A1_long_format},\eqref{A2_long_format}, \eqref{A3_long_format} are equivalent to \eqref{obj_factors_simplified}, \eqref{An_efficient_format}  by knowing the following properties:
 \begin{enumerate} [leftmargin=2cm]
    \item $KL(\M{A}|| \M{B} ) = KL(\text{vec}(\M{A})|| \text{vec}(\M{B}))$
    \item  $KL(\M{A}|| \M{B} ) = KL(\M{A}^T|| \M{B}^T )$
    \item $ (\M{E}^T \otimes \M{C}) \text{vec}(\M{D})=\text{vec}(\M{C} \M{D} \M{E} )$  \quad \cite{neudecker1969some}
    \item $\text{vec}(\M{A}_3 \odot \M{A}_2) = ((\M{I}_R \odot \M{A}_3) \otimes \M{I}_{I_2})  \text{vec}(\M{A}_2)$ \quad \cite{roemer2012advanced}.
     \item $\text{vec}(\M{A}_2 \odot \M{A}_1) = \Big( \M{I}_{I_2R} \odot (\M{A}_1(\M{I}_R \otimes \V{1}_{1 \times I_2})) \Big) \text{vec}(\M{A}_2)$ \quad \cite{roemer2012advanced}.
     \item $\M{I}_{I_1I_2} = (\M{I}_{I_1} \otimes \M{I}_{I_2})= (\M{I}_{I_2} \otimes \M{I}_{I_1}) =\M{I}_{I_2I_1} $  \quad \cite{neudecker1969some}.
\end{enumerate}

\par
\begin{figure}[h!]
    \begin{subfigure}[]{.32\textwidth}
        \centering\includegraphics[height=1.5in]{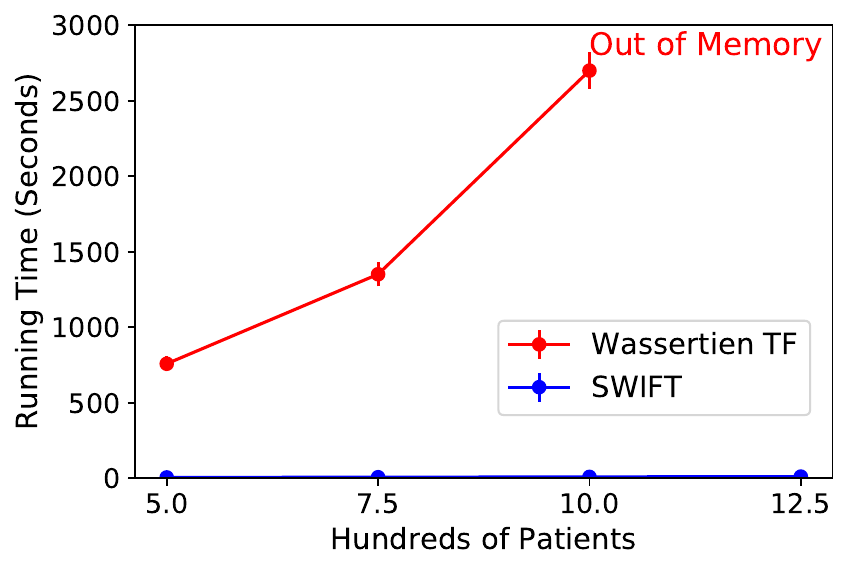}
        \caption{Time vs Patient Mode on Sutter Data.}\label{fig:Sutter_dim}%
    \end{subfigure}%
    \begin{subfigure}[]{.32\textwidth}
        \centering\includegraphics[height=1.5in]{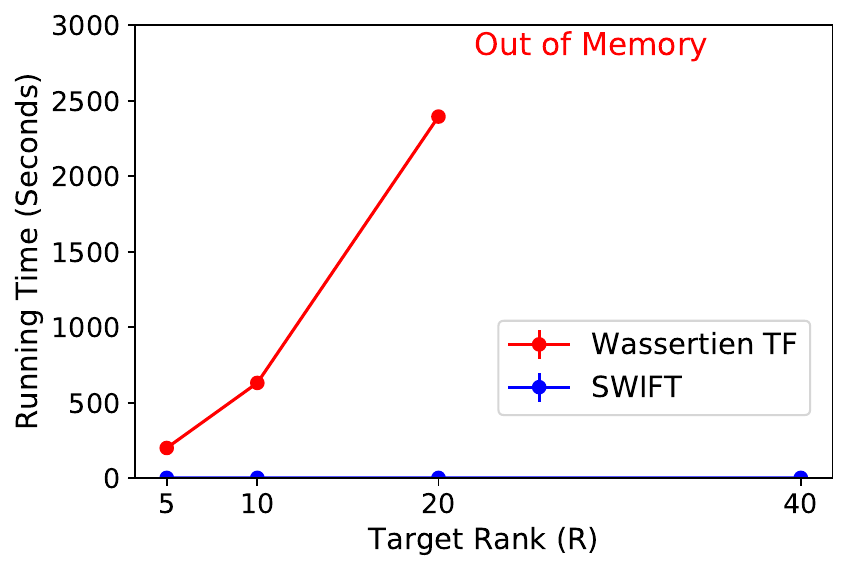}
        \caption{Time vs R on Sutter Data.}\label{fig:Sutter_R}
    \end{subfigure}%
    \begin{subfigure}[]{.32\textwidth}
        \centering\includegraphics[height=1.5in]{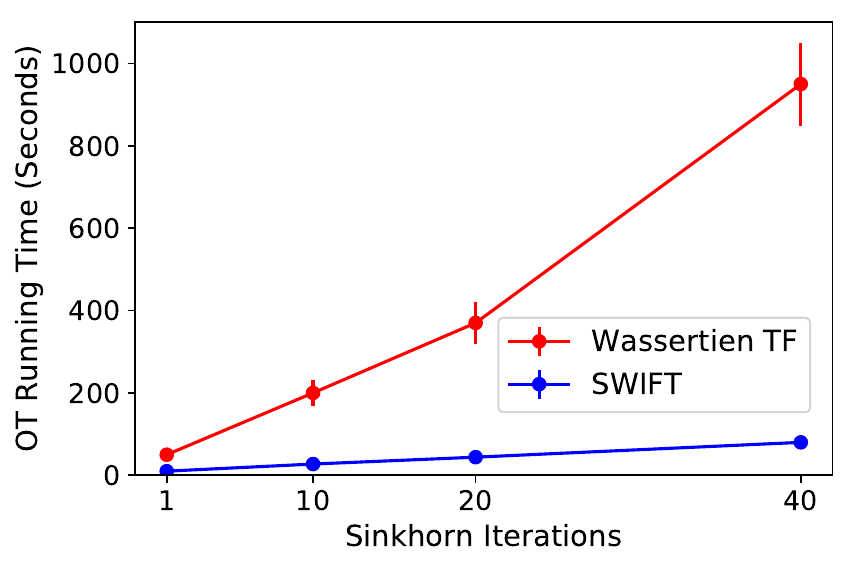}
        \caption{Time vs Sinkhorn on Sutter Data.}\label{fig:Sutter_sinkhorn}
    \end{subfigure}\\[5mm]
    \begin{subfigure}[]{.32\textwidth}
       \centering \includegraphics[height=1.5in]{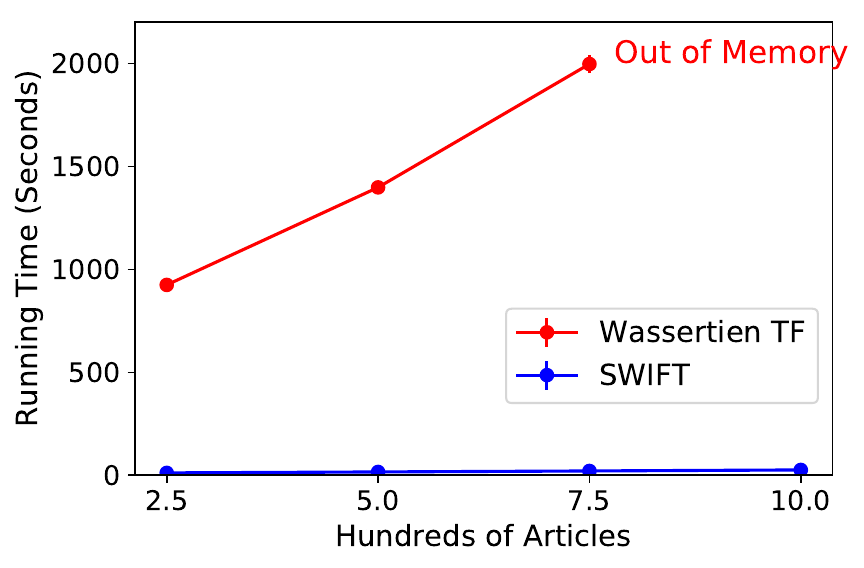}
       \caption{ Time vs Article Mode on BBC News.} \label{fig:BBC_dim}%
     \end{subfigure}%
    \begin{subfigure}[]{.32\textwidth}
        \centering\includegraphics[height=1.5in]{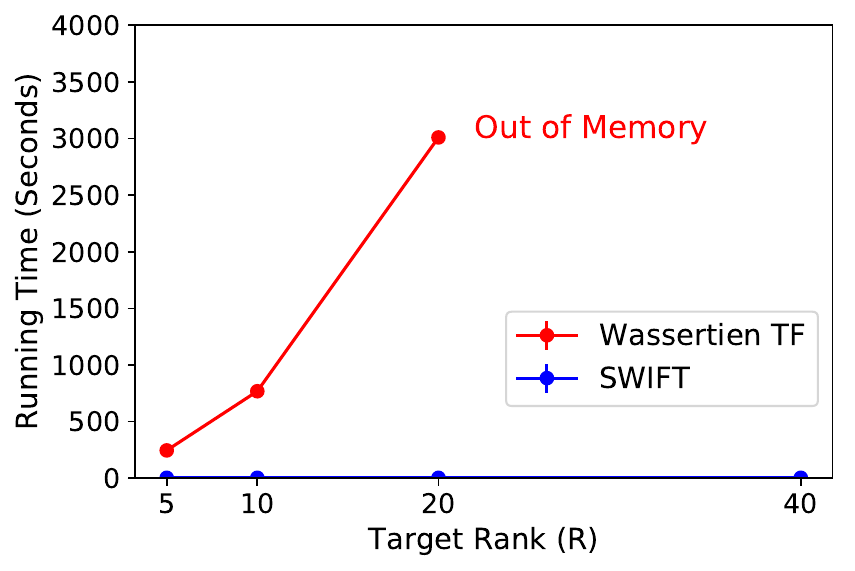}
        \caption{Time vs R  on BBC News.}\label{fig:BBC_R}
    \end{subfigure}%
    \begin{subfigure}[]{.32\textwidth}
        \centering\includegraphics[height=1.5in]{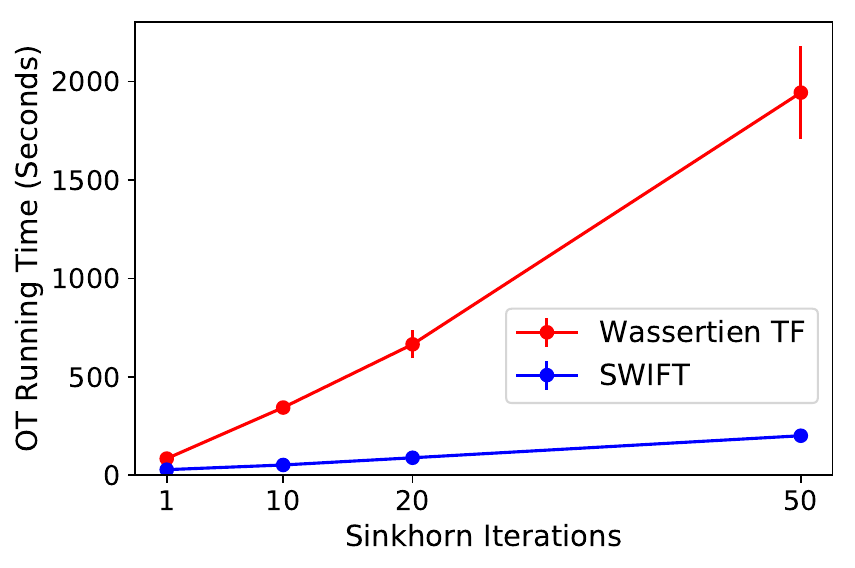}
        \caption{Time vs Sinkhorn on BBC News.}\label{fig:BBC_sinkhorn}
\end{subfigure}
\caption{Average and standard deviation of running time in seconds for one iteration (as an average of five) on Sutter and  BBC NEWS Data sets by increasing 1) a mode size 2) target rank (R) 3) Sinkhorn iteration.  For Figures~\ref{fig:Sutter_sinkhorn}, \ref{fig:BBC_sinkhorn} we report OT running time and for the rest we present the total running time for one iteration. In Figure~\ref{fig:Sutter_dim} for 1250 patients and Figure~\ref{fig:Sutter_R} for $R=40$, Figure~\ref{fig:BBC_dim} for 1000 articles, and Figure~\ref{fig:BBC_R}, execution in Wasserstein TF (direct baseline) failed due to the excessive amount of memory request.  }
\label{scalability_swift_naive_baseline}
\end{figure}

 \textbf{Results:} 
In Figure~\ref{scalability_swift_naive_baseline}, we vary 1) size of a mode, 2) target rank (R), and 3) number of Sinkhorn iterations in optimal transport problems and report the average and standard deviation of running time (in seconds) for one iteration as an average of five different runs for Sutter and BBC NEWS data sets.

Figures \ref{fig:Sutter_dim} depicts the time as we increase the number of patients ($\{ 500,750, 1000,1250 \}$). \mname is up to $293 \times$ faster than the Wasserstein TF model. Execution in the direct Wasserstein TF failed for a tensor of $1250 \times 100 \times 100$, due to the excessive amount of memory needed.  Next, we compare the scalability of both methods by varying the value for R on Figure \ref{fig:Sutter_R}. \mname achieves up to $928 \times$ faster computation. Figure \ref{fig:Sutter_sinkhorn} depicts the optimal transport time\footnote{We avoid updating factor matrices here and just record the running time of optimal transport problem.} for \mname and Wasserstein TF for a tensor with size $1500 \times 500 \times 500$. Generally, more Sinkhorn iterations lead to a better solution for the optimal transport problems. We increase the value of Sinkhorn iterations and \mname achieves up to $12 \times$ speed up over Wasserstein TF suggests that speedup strategies introduced in Section~\ref{OT_sol} are beneficial.

Next, we perform the experiments for BBC NEWS data.  Figure \ref{fig:BBC_dim} demonstrates the running time for one iteration as we increase the number of articles in BBC NEWS data ($\{ 500,750, 1000,1250 \}$).  \mname is up to $ 97 \times$ faster than the direct Wasserstien tensor factorization model in Figure~\ref{fig:BBC_dim}. Same as before, execution in the direct Wasserstien TF failed  for a tensor of $1250 \times 100 \times 100$,  due to the excessive amount of memory.   We measure the scalability of both methods by varying the value for R ($R=\{5,10,20,40 \}$) on Figure \ref{fig:BBC_R} where \mname achieves  up to $ 718 \times$ faster computation. Again for $R=40$, direct Wasserstien TF failed to execute due to out of memory. Finally,  Figure  \ref{fig:BBC_sinkhorn} represents the optimal transport running time\footnotemark[\value{footnote}] for one iteration (as an average of 5) for \mname and Wasserstien tensor factorization for a tensor with size $1500 \times 500 \times 500$.  We increase the  value of Sinkhorn iterations ($\{ 1, 10, 20, 50 \}$).  \mname achieves up to $10 \times$  speed up over Wasserstien tensor factorization. 

Figure~\ref{scalability_swift_naive_baseline} suggests the strategies we introduced in Sections~\ref{OT_sol} and \ref{Fac_mat_sol} are beneficial and significantly reduce the running time of \mname in compare to the direct implementation.

 \subsection{Effect of Cost Matrix on Classification Performance} \label{appendix:cost_mat:effect}
 In order to understand the power of cost matrix (\M{C}) on classification performance, we  compare  different cost matrices including the one we introduced in section~\ref{COST_MAT_CALC},  random and 1-identity (matrix of ones with a zero diagonal) cost matrices for both BBC NEWS and Sutter datasets.  When the cost matrix is 1-identity then Wasserstein distance can be seen as total variation distance \cite{cuturi2014ground}.

From Table~\ref{tab:swift_diff_cost} we summarize our observations as follow : 1) using random cost matrices  drop the performance. 2) 1-identity cost matrix performs better than random cost matrices but still not as good as cost matrices introduced in Section~\eqref{COST_MAT_CALC}. As you can see \mname with  1-identity cost matrices have better performance than other well-known CP baselines (Comparing with Table~\ref{tab:pred_power_datasets}).

\begin{table*}[h]
  \centering
  \caption{ Average and standard deviation of accuracy on BBC NEWS and PR-AUC score  on Sutter data sets by considering different cost matrices for different values of target-rank ($R=[5,10,20,30,40]$).}
    \begin{tabular}{clrrrrr}
          &       & \multicolumn{1}{c}{R=5} & \multicolumn{1}{c}{R=10} & \multicolumn{1}{c}{R=20} & \multicolumn{1}{c}{R=30} & \multicolumn{1}{c}{R=40}  \\
    \toprule
    \multirow{3}[2]{*}{\makecell{\textbf{BBC} \\ \textbf{NEWS} \\ \textbf{Data}}} 
          & \mname (Random) & .621 $\pm$ .041  & .693 $\pm$ .053 & .684 $\pm$  .038 & .712 $\pm$ .047   &  .684 $\pm$ .022  \\[1pt]
          & \mname (\M{1}-Identity)  & .709 $\pm$ .016  & .775 $\pm$ .019 & .793  $\pm$ .032 & .775 $\pm$ .023  & .771 $\pm$ .013 \\[1pt]
          & \mname &  \textbf{.759 $\pm$ .013} &  \textbf{.781 $\pm$ .013}  & \textbf{.803 $\pm$ .010} &    \textbf{.815 $\pm$  .005} & \textbf{.818  $\pm$ .022} \\[1pt]
    \midrule
    \multirow{3}[2]{*}{\makecell{\textbf{Sutter} \\ \textbf{Data}}}  
          & \mname (Random) & .340 $\pm$ .103  & .347 $\pm$ .091 & .341 $\pm$ .083 & .352 $\pm$ .041 & .354 $\pm$ .074 \\[1pt]
          & \mname (\M{1}-Identity)  & .363 $\pm$ .105  & .349 $\pm$ .079 & .344 $\pm$ .073 & .358 $\pm$ .063 & .368 $\pm$ .074 \\[1pt]
          & \mname & \textbf{.364 $\pm$ .063}  & \textbf{.350 $\pm$ .031} & \textbf{.350 $\pm$ .040 }& \textbf{.369 $\pm$ .066} & \textbf{.374 $\pm$ .044} \\[1pt]
        
    \bottomrule
    \end{tabular}%
  \label{tab:swift_diff_cost}%
\end{table*}%

\subsection{More Results on HF Phenotyping} \label{apndx_hf_pheno}
We show the remaining phenotypes positively associated with HF in Table~\ref{app:tab:phenos}.

\begin{table}[h!]\small
  \centering
  \caption{The phenotypes positively associated with HF. All phenotypes are annotated and endorsed by a medical expert. The unlisted ones are either negatively or not associated with HF.}
    \begin{tabular}{ll}
    \toprule
    \textbf{HF with long-term Diabetes (Weight= 14.62)} & \textbf{HF with Dysrhythmias (Weight= 13.73)} \\
    \midrule
    \textcolor{red}{Dx-Diabetes with ketoacidosis or uncontrolled diabetes} & \textcolor{red}{Dx-Cardiac dysrhythmias [106.]} \\
    \textcolor{red}{Dx-Other mycoses} & \textcolor{red}{Dx-Diabetes mellitus without complication [49.]} \\
    \textcolor{red}{Dx-Diabetes mellitus without complication [49.]} & \textcolor{red}{Dx-Other viral infections [7.]} \\
    \textcolor{blue}{Rx-Corticosteroids - Topical} & \textcolor{blue}{Rx-Coumarin Anticoagulants} \\
    \textcolor{blue}{Rx-Biguanides} & \textcolor{blue}{Rx-Gout Agents} \\
    \textcolor{blue}{Rx-Central Muscle Relaxants} & \textcolor{blue}{Rx-Calcium Channel Blockers} \\
    \toprule
    \textbf{CAD related HF (Weight= 10.04)} & \textbf{Aging and frail related HF (Weight= 9.72)} \\
    \midrule
    \textcolor{red}{Dx-Cardiac dysrhythmias [106.]} & \textcolor{red}{Dx-Urinary tract infections [159.]} \\
    \textcolor{red}{Dx-Acute cerebrovascular disease [109.]} & \textcolor{red}{Dx-Genitourinary symptoms and ill-defined conditions [163.]} \\
    \textcolor{red}{Dx-Disorders of lipid metabolism [53.]} & \textcolor{red}{Dx-Other bone disease and musculoskeletal deformities [212.]} \\
    \textcolor{blue}{Rx-Coumarin Anticoagulants} & \textcolor{blue}{Rx-Urinary Anti-infectives} \\
    \textcolor{blue}{Rx-Oil Soluble Vitamins} & \textcolor{blue}{Rx-Fluoroquinolones} \\
    \textcolor{blue}{Rx-HMG CoA Reductase Inhibitors} & \textcolor{blue}{Rx-Anti-infective Misc. - Combinations} \\
    \toprule
    \textbf{HF with Pulmonary disease (Weight= 9.58)} & \textbf{ HF with air pathway blockage (Weight= 8.98)} \\
    \midrule
    \textcolor{red}{Dx-Other and unspecified asthma} & \textcolor{red}{Dx-Essential hypertension [98.]} \\
    \textcolor{red}{Dx-Chronic airway obstruction; not otherwise specified} & \textcolor{red}{Dx-Allergic reactions [253.]} \\
    \textcolor{red}{Dx-Acute bronchitis [125.]} & \textcolor{red}{Dx-Esophageal disorders [138.]} \\
    \textcolor{blue}{Rx-Sympathomimetics} & \textcolor{blue}{Rx-Calcium Channel Blockers} \\
    \textcolor{blue}{Rx-Fluoroquinolones} & \textcolor{blue}{Rx-Beta Blockers Cardio-Selective} \\
    \textcolor{blue}{Rx-Opioid Combinations} & \textcolor{blue}{Rx-Proton Pump Inhibitors} \\
    \toprule
    \textbf{HF with systematic inflammation    (Weight= 6.45)} & \textbf{HF with thyroid dysfunction (Weight= 4.08)} \\
    \midrule
    \textcolor{red}{Dx-Essential hypertension [98.]} & \textcolor{red}{Dx-Immunizations and screening for infectious disease [10.]} \\
    \textcolor{red}{Dx-Chronic kidney disease [158.]} & \textcolor{red}{Dx-Essential hypertension [98.]} \\
    \textcolor{red}{Dx-Disorders of lipid metabolism [53.]} & \textcolor{red}{Dx-Other thyroid disorders} \\
    \textcolor{blue}{Rx-ACE Inhibitors} & \textcolor{blue}{Rx-Antihypertensive Combinations} \\
    \textcolor{blue}{Rx-HMG CoA Reductase Inhibitors} & \textcolor{blue}{Rx-Impotence Agents} \\
    \textcolor{blue}{Rx-Calcium Channel Blockers} & \textcolor{blue}{Rx-Angiotensin II Receptor Antagonists} \\
    \bottomrule
    \end{tabular}%
  \label{app:tab:phenos}%
\end{table}%

\end{document}